\documentclass[11pt]{article}

\date{}

\usepackage[utf8]{inputenc} 
\usepackage[T1]{fontenc}    
\usepackage[colorlinks=true,linkcolor=blue,allcolors=blue]{hyperref}       
\usepackage{url}            
\usepackage{booktabs}       
\usepackage{amsfonts}       
\usepackage{nicefrac}       
\usepackage{microtype,nicefrac}      
\usepackage{xspace}
\usepackage{natbib}

\usepackage{comment}
\usepackage{amsmath}
\usepackage{amsthm}
\usepackage{graphicx,subcaption}
\usepackage{rotating}

\usepackage{amssymb}
\usepackage{autobreak}
\usepackage{mathtools}
\usepackage{xcolor}
\usepackage{bbm}
\usepackage[ruled,vlined]{algorithm2e}
\usepackage{algorithmic}

\makeatletter

\DeclareMathOperator*{\argmin}{arg\,min}
\DeclareMathOperator*{\esssup}{ess\,sup}

\newcommand{\barkernel}{\bar\kappa}

\newcommand{\Eone}{\textbf{E1}}
\newcommand{\Etwo}{\textbf{E2}}
\newcommand{\Ethree}{\textbf{E3}}
\newcommand{\Efour}{\textbf{E4}}

\newcommand{\kernel}{\kappa}
\newcommand{\FuncClass}{\mathbb{F}}
\newcommand{\Rade}{\mathcal{R}}
\newcommand{\loss}{{L}}
\newcommand{\impW}{\omega}

\newcommand{\imptheta}{\theta}
\newcommand{\impthetaV}{\theta}
\newcommand{\DS}{D_S}
\newcommand{\DT}{D_T}
\newcommand{\impV}{\omega}
\newcommand{\VecOne}{\Vec{1}}
\newcommand{\FunOne}{\mathbbm{1}}
\newcommand{\Dpg}{\Delta_{p_g}}
\newcommand{\Dqg}{\Delta_{q_g}}
\newcommand{\DTg}{\Delta_{T_g}}
\newcommand{\Dpu}{\Delta_{p_u}}
\newcommand{\Dqu}{\Delta_{q_u}}
\newcommand{\DTu}{\Delta_{\T_u}}
\newcommand{\measures}{\mathcal{P}}
\newcommand{\gmax}{g_{\max}}
\newcommand{\umax}{u_{\max}}
\newcommand{\impthetamax}{\imptheta_{\max}}
\newcommand{\G}{\mathcal{G}}

\newcommand{\Comp}{\mathbb{C}}
\newcommand{\HS}{\mathcal{HS}}
\newcommand{\Hilbert}{\mathcal{H}}

\newcommand{\dinf}{d_{\infty}}

\newcommand{\ERM}{\textsc{\small{ERM}}\xspace}

\renewcommand{\Re}{\mathbb{R}}
\newcommand{\Natural}{\mathbb{N}}

\newcommand{\F}{\mathcal F}

\newcommand{\X}{\mathcal X}
\newcommand{\T}{\mathcal T}

\renewcommand{\L}{\mathcal L}

\newcommand{\Y}{\mathcal Y}
\newcommand{\V}{\mathcal V}

\newcommand{\E}{\mathbb E}

\newcommand{\Prob}{\mathbb P}
\newcommand{\Qrob}{\mathbb Q}

\newcommand{\wh}{\widehat}
\newcommand{\wb}{\overline}

\newtheorem{lemma}{Lemma}[section]
\newtheorem{remark}{Remark}[section]

\newtheorem{proposition}{Proposition}[section]

\newtheorem{theorem}{Theorem}[section]

\title{Importance Weight Estimation and  Generalization in Domain Adaptation under Label Shift}

\author{%
  Kamyar Azizzadenesheli\\ 
  Department of Computer Science\\
  Purdue University\\
  \texttt{kamyar@purdue.edu}\thanks{A version of this manuscript is published at the IEEE Transactions on Pattern Analysis and Machine Intelligence} \\
}

\begin{document}
\maketitle

\begin{abstract}
We study generalization under labeled shift for categorical and general normed label spaces. We propose a series of methods to estimate the importance weights from labeled source to unlabeled target domain and provide confidence bounds for these estimators. We deploy these estimators and provide generalization bounds in the unlabeled target domain.

\end{abstract}
\textit{keywords:
  label shift, domain adaptation, integral operator, generalization.
}

\section{Introduction}
In supervised learning, we usually make predictions on an unlabeled target set by training a predictive model on a labeled source set. This is a sensible approach, especially when the source and target sets are expected to have i.i.d. samples from the same distribution. However, this assumption does not hold in many real-world applications. For example, oftentimes, predictions for medical diagnostics are based on data from a particular population (or the same population in a few years back) of patients and machines due to the limited variety of data that is available in the training phase. Let us consider the following hypothetical scenario. A classifier is trained to predict whether a patient has contracted a severe disease in country A based on the physical measurements and diagnosis (labeled source data) in country A. The disease is potentially even more prevalent in country B, where no experts are available to make a reliable diagnosis. Ergo, the following questions could be relevant in this scenario:
\begin{itemize}
\item  Suppose we send a group of volunteers to country B with a preliminary set of equipment to record measurements of patients' symptoms. How can the data in country A and the new unlabeled data from country B be used to update the classifier to give good diagnostic predictions for patients in country B?

\item What if, along with the few diseases we are considering, we are also interested in predicting disease levels where the task is to predict continuous labels, therefore regression?
\end{itemize}

Similar problems arise in many other domains besides health care, but we will use the medical example throughout the paper for ease of presentation. In the above example, country A denotes the source domain, and country B denotes the target domain. In this example, the distribution of covariates-label in the source may differ from that of the target domain. Therefore, a predictor learned on the source domain might not carry on to the target domain. In the above hypothetical example, consider the case that given a person having a disease $y$, the distribution of their symptoms is similar in both source and target domains, but the proportion of people having different diseases differs.  

Consider another example where the source domain is data of a city in Fall, and the target is the data of the same city, a few months later, in Spring. The disease distribution in Spring may differ from that of Fall, but given a person caught a disease, that person expresses similar symptoms in both seasons. In domain adaptation, the label shift is a class of problems where there is a shift in label distribution from source to the target domain, but the conditional distribution of covariates stays unchanged. 

We used the medical example to motivate the setting. However, the label shift paradigm is a common in practice. For example, one can use image data of a population to train a model that later is required to adapt to the different demographic environment. In scientific application, to tackle inverse problems, e.g., sensing, or partial differential equations, one may learn an inverse map that maps an outcome to a source. However, later the model needs to be adapted to new distributions of sources in different applications. 

In a generic label shift domain adaptation problem, we are given a set of labeled samples from the source domain but unlabeled samples from the target domain. In such settings, the task is to learn a model with low loss in the target domain. In this paper, we study two settings, $(i)$ categorical label spaces, $(ii)$ general normed label spaces. We propose a series of methods to estimate the importance weight from source to target domain for both of these settings. Prior works study categorical label spaces and propose to use a label classifier to estimate the importance weight vector, later used in empirical risk minimization (\ERM), to learn a good classifier. \citet{lipton2018detecting} provides a method to estimate the importance weight vector and guarantee estimation error under strong assumptions that the error in the estimation of confusion matrix should be full rank, the confusion matrix should be a square matrix, the number of samples should be larger than an unknown quantity. \citet{azizzadenesheli2019regularized} propose another estimator and relax the assumptions in the prior work, but still based on label classifier. In this paper, we first relax the requirement to use a classifier to estimate the importance weight vector. This step improves the conditioning on the inverse operator (inverse of the transition operator from labels to predicted statistics) and exploits the spectrum of the forward operator appropriately. We propose a regularized estimator, which is an extension to \citet{azizzadenesheli2019regularized}.

We further improve the analysis in \citet{lipton2018detecting} and propose a estimator using general functions (rather than being limited to classifier). For this estimator, we relax the requirement that the confusion matrix should be a square matrix and further relax the strong assumption that the error matrix is full rank. Note that, this analysis also applies to the case where classifiers are used for importance weight estimation and is considered as an improvement to \citet{lipton2018detecting}. Moreover, this analysis results in an estimation bound as tight as the one for the regularized estimator. 

For the case of $(ii)$ general normed label spaces, we propose two estimators to estimate the importance weight functions. The first estimator is based on the traditional approach used in inverse problems of compact operators, which directly estimates the inverse operator\citep{kress1989linear}. Such an estimator requires the number of samples to be larger than an unknown quantity, which can be considered as a limitation of the traditional methods in inverse problems. The second approach is a novel approach based on regularized optimization in Hilbert spaces, and the derived bound holds for any number of samples. Therefore this approach has an advantage over the traditional approaches and has its own independent importance beyond domain adaptation. As mentioned before, such problems are common in partial differential equations and reinforcement learning.

For both of the proposed estimators, we provide concentration bound on our estimate based on a desirable norm and show that the estimation errors vanish as the number of samples grows. Finally, we deploy these importance weight estimates in importance weighted \ERM and provide generalization guarantees of \ERM in these settings.  Using the existing generalization analysis based on Rademacher complexity, R\'enyi divergence, and MDFR Lemma in \citet{azizzadenesheli2019regularized}, we show the generalization property of the importance weighted empirical risk minimization on the unseen target domain for both categorical and general normed label spaces.

\section{Preliminaries}
Let $\X$ and $\Y$ be two sets, denoting the sets of covariates and labels, and $\measures$ denote a space of measures on a measure space $(\X\times\Y,\F)$. For the product space of $\X\times\Y$, let $(\X\times\Y,\F,\Prob)$ denote the probability space on the source domain, and respectively, $(\X\times\Y,\F,\Qrob)$ denote the probability space on the target domain where $\Prob,\Qrob\in \measures$. For the label space $\Y$, let $\F_\Y\subset\F$, generated by subsets of $\Y$, denote the sub-$\sigma$-algebra of $\F$, with $\Prob_\Y$ and $\Qrob_\Y$ as restrictions of $\Prob$ and $\Qrob$ to $\F_{\Y}$. Similarly for $\X$, i.e., $\Prob_\X$ and $\Qrob_\X$ on $\F_{\X}$.  We consider the setting where $\Qrob_\Y$ is absolutely continuous with respect to $\Prob_\Y$. We define the shift between source and target domains as a label shift type, if for any random variable $R:\X\times\Y\rightarrow \Re$, integrable under $\Prob$, there exist a version for $\E_\Prob[R|\F_\Y]$ which is equal to a version of $\E_\Qrob[R|\F_\Y]$ almost surely.

When $\X$ and $\Y$ are normed space, for a given complex separable Hilbert space $\wb\Hilbert$ accompanied with inner product $\langle\cdot,\cdot\rangle$ and norm $\|\cdot\|$, let $\T$ denote a linear bounded operator $\T:\wb\Hilbert\rightarrow \wb\Hilbert$. We say an operator $\T$ is Hilbert-Schmidt if $\sum_{i\geq1}\|\T e_i\|<\infty$ for any set of bases $e_i$ of $\wb\Hilbert$. Note that, the separability of $\wb\Hilbert$ is required to have countable bases. The space of Hilbert-Schmidt operators ($\HS$) is also a Hilbert space under inner product defined as $\langle\T_1,\T_2\rangle_\HS = \sum_i\langle\T_1e_i,\T_2e_i\rangle$ for $\T_1,\T_2\in\HS$, and respectively $\|\cdot\|_\HS$ as the corresponding norm \citep{lang2012real,kress1989linear}. We define a reproducing kernel Hilbert space (RKHS) $\Hilbert$, a Hilbert space of functions $h:\Y\rightarrow\Comp$, accompanied with a symmetric positive definite continuous reproducing kernel $\kernel:\Y\times\Y\rightarrow\Comp$, such that $\barkernel:=\sup_{y\in\Y}\kernel(y,y)$ is finite.

In this paper, in order to account for the shifts between source and target domains, we consider the exponent of the infinite and second order R\'{e}nyi divergences, the notions deployed also in prior works~\citep{cortes2010learning,azizzadenesheli2019regularized}. For the restricted probabilities measures $\Prob_\Y$ and $\Qrob_\Y$, we have:
\begin{align*}
    \dinf(\Qrob_\Y||\Prob_\Y):=\esssup \frac{d\Qrob_\Y}{d\Prob_\Y},~ d(\Qrob_\Y||\Prob_\Y):=\E_\Prob\left[\left(\frac{d\Qrob_\Y}{d\Prob_\Y}\right)^2\right].
\end{align*}
where $\frac{d\Qrob_\Y}{d\Prob_\Y}\in L^1(\Prob_\Y)$, the importance weight function, is the Radon–Nikodym change of measure function, and $\E_\Prob$ and $\E_\Qrob$ denote expectation with respect to $\Prob$ and $\Qrob$ respectively. To simplify notation, let $\impW=\frac{d\Qrob_\Y}{d\Prob_\Y}$, denote the importance weight function. 


In label shift setting, we are interested in the prediction task in the target domain. In other words, for a given loss function $\ell:\Y\times\Y\rightarrow [0,1]$, and a function class $\FuncClass$, we are interested in finding a function $f\in\FuncClass$, $f:\X\rightarrow\Y$ with small expected loss, $\loss(f,\Qrob) = \E_{\Qrob}[\ell(Y,f(X))]$. We note that,
\begin{align*}
    \loss(f,\Qrob) = \E_{\Qrob}\left[\ell(Y,f(X))\right] =  \E_{\Prob}\left[\impW(Y) \ell(Y,f(X))\right].
\end{align*}

\begin{table}[t]
\centering
\caption{Label shift domain adaptation}
 \begin{tabular}{l| c } 
 \toprule
 \textbf{Problem setting} & \textbf{Data regime} \\ 
 \midrule
 \shortstack{1) $\Prob_\Y$ and $\Qrob_\Y$ can be different\\ ~} & \shortstack{ Data set $\DS=\lbrace x_i,y_i\rbrace_{i=1}^n$}\\
\shortstack{2) For any random variable $R$,\\
  \quad $\E_\Prob[R|\F_\Y]=\E_\Qrob[R|\F_\Y]$ a.s.}
  & \shortstack{ Data set $\DT=\lbrace x_i\rbrace_{i=1}^m$}\\
 \bottomrule
\end{tabular}
\label{table:Setting}
\end{table}

In the label shift setting, we have access to $n$ labeled data points from the source domain $\DS=\lbrace x_i,y_i\rbrace_{i=1}^n$, but $m$ unlabeled samples form the target domain $\DT=\lbrace x_i\rbrace_{i=1}^m$ (see Table~\ref{table:Setting}). In the case where we are provided with the importance weight function $\impW$, we could deploy importance weighted \ERM on the source domain, 
\begin{align*}
     \wh f\in\argmin_{f\in\FuncClass}\loss(f,\wh\Prob_n,\impW):=\E_{\wh\Prob_n}\left[\impW(Y) \ell(Y,f(X))\right].
\end{align*}
where $\wh\Prob$ is the empirical measure induced by $\DS$, to obtain a prediction function $\wh f$. However, in practice, we do not have access to $\impW$, and need to estimate it using the knowledge of $\DS$ and $\DT$. Having access to the data set $\DS$ of size $n$, we split the data set to two subsets. We randomly select $\alpha$ portion of the $\DS$, and use these $\alpha n $ samples and its corresponding empirical measure $\wh\Prob_{\alpha n}$, along with samples in $\DT$ to  estimate importance weight $\wh\impW$, and utilize the remaining $(1-\alpha)n$ samples of $\DS$ for the importance weighted \ERM to obtain $\wh f$, i.e., 
\begin{align}\label{eq:riskminimizer}
     \!\!\!\!\!\!\wh f\in\argmin_{f\in\FuncClass}\loss(f,\wh\Prob_{(1-\alpha)n},\wh\impW)\!:=\!\E_{\wh\Prob_{(1-\alpha) n}}\!\left[\wh\impW(Y) \ell(Y,f(X))\right]\!.\!\!\!\!\!
\end{align}
We propose a series of methods to estimate $\wh\impW$, up to their confidence bounds, and provide generalization guarantees for the importance weighted  \ERM 
of $\loss(f,\wh\Prob_{(1-\alpha)n},\wh\impW)$.

\section{Problem Setup}
We study estimation of importance weight functions in both classification and regression settings. 

\subsection{Categorical Label Spaces}

Consider the task of multi class classification where the task is to predict a class given covariates. The label space $\Y=[k]=\lbrace 1,2,\ldots, k\rbrace$ where $k$ is the number of classes\footnote{For simplicity we describe the results for finite $k$, but the standard generalization of norms to countably infinite sets extends the results in this paper to general countable label sets.}. In this setting, we represent the importance weight function as a $k$ dimensional vector $w\in\Re^k$ such that $\impV_i=\impW(i)$ for $i\in\Y$. For any dimension $d$ and measurable function $g:\X\rightarrow [-1,1]^d$, we show how $w$ relates the $p_g:=E_\Prob[g]$ and $q_g:=E_\Qrob[g]$. Note that, to compute these expectations we do not need the associate labels. In the following we denote $\gmax=\|\sup_{\mu\in\measures}\int_{\X\times\Y}g(X)d\mu(X,Y)\|$. Using the rules of conditional expectations, we have,
\begin{align}\label{eq:finite_confusion}
    q_g=E_\Qrob[g]&=E_\Qrob[E_\Qrob[g|\F_\Y]]\nonumber\\
    &=E_\Qrob[E_\Prob[g|\F_\Y]]\nonumber\\
    &=E_\Prob[\impW E_\Prob[g|\F_\Y]]
\end{align}
Let $T_g:\Re^k\rightarrow\Re^d$ denote the corresponding linear (matrix) operator in Eq.~\ref{eq:finite_confusion}, i.e., $q_g=T_g\impV$, represented as $(T_g)_{i,j}=E_\Prob[g_i|\F_\Y](j)\Prob(Y=j)$. 

\begin{sloppypar}
\begin{remark}
When $g\in\FuncClass$ is a classifier, i.e., outputs a class with a one-hot encoding representation, $(T_g)_{i,j} = \Prob(g(X)=i,Y=j)$. This special case of $g$, has been previously observed by \citet{lipton2018detecting,azizzadenesheli2019regularized} in Eq.~\ref{eq:finite_confusion}. The general form of Eq.~\ref{eq:finite_confusion} is one of the contribution of the present paper. 
\end{remark}
\end{sloppypar}

In Eq.~\ref{eq:finite_confusion}, having access to $q_g$ and well conditioned (full column rank) $T_g$, one can compute $\impV$ using Moore–Penrose inverse of $T_g$, i.e., $T_g^\dagger$, and we have ${\impV} := T_g^\dagger q_g$.
In the case where $\Prob = \Qrob$ a.s., a vector of all ones is a feasible $\impV$ in $q_g=T_g\impV$, i.e., importance weights are all one. When approximating $\impV$ in the presence of no additional side information about $\Prob$ and $\Qrob$, a homogeneous regularization around vector of all ones is desirable, i.e., finding $\impV$ satisfying  $q_g=T_g\impV$ and is closest to vector of all one. However, 
$T_g^\dagger q_g$ outputs a importance vector which satisfies $q_g=T_g\impV$, but is closest to the zero vector in $L_2$-norm sense. Therefore, even in the case of no shift, using $T_g^\dagger q_g$ may result in estimates of importance weight with many entries equal to zero, resulting in undesirable importance weighted \ERM classifier.

\textbf{Reformulation technique:}
Instead of solving for $\impV$ in $q_g=T_g\impV$, we solve for $\impthetaV$ where $\impV=\VecOne+\impthetaV$, $\VecOne$ is a $k$ dimensional vector of ones, and $\VecOne,\impthetaV\in\Re^k$. Note that, when $\Prob = \Qrob$ a.s., a vector of all zeros is a feasible $\impthetaV$, i.e., importance weights are all one. Using $\impthetaV$ formulation, we have:
\begin{align}\label{eq:eq_vector}
    q_g-p_q= T_g\impthetaV 
\end{align}
where the right hand side is equal to zero when $\Prob = \Qrob$ a.s. . It is important to note that $\imptheta$ accounts for the amount of shift. Small in value $\imptheta$ denotes small shifts, while large $\imptheta$ represent large shifts, and potentially harder problems. We denote $\imptheta_{\max}$ as the maximum norm on allowed shifts. One can find a desirable $\impthetaV$ using, $\impthetaV := T_g^\dagger (q_g-p_g)$ which is a $\impthetaV$ solution to $q_g-p_q = T_g\impthetaV$ and is closest to vector of all zeros (origin) in $L_2$-norm sense. In this paper, we focus on estimating the importance weight through estimating $\impthetaV$.

\subsection{General Normed Label Spaces}

In the following, we consider the case where the label space is a normed vector space. Let $\Hilbert$ denote the RKHS defined in the preliminaries section. For a given function $u:\X\rightarrow \Y$, we define $\V_u:\measures\rightarrow\Hilbert$, an operator such that $\V_u(\mu)= \int_{\X\times\Y} k_{u(x)}d\mu(x,y)$ for any measure $\mu\in\measures$. In other words, $\forall y'\in\Y$,  $(\V_u \mu)(y')= \int_{\X\times\Y} k(y',u(x))d\mu(x,y)$. Let, $\forall y\in\Y$, $\kernel^u_y\in\Hilbert$ denote a version of $\E_\Prob[\kernel_{u(X)}|\F_\Y](y)$. We define an integral operator by a reproducing kernel and a positive measure $\Prob_\Y$ on $(\Y,\F_Y\!)$, as an operator $\T_u\!:\!\Hilbert\rightarrow\!\!\Hilbert$ such that,
\begin{align*}
    \T_u:=\int_\Y(\kernel^u_y\otimes\kernel_y)d\Prob_\Y(y),
\end{align*}
denote general integral operators from $\Hilbert$ to $\Hilbert$. Note that, for any $h\in\Hilbert$, we have $\T_u h:=\int_\Y\kernel^u_y\langle\kernel_y,h\rangle d\Prob_\Y(y).$ We define $q_u\in\Hilbert$ as $q_u:=\V_u(\Qrob)$. Let $\FunOne\in\L^1$ denote a constant function with value $1$, and $p_u=\V_u(\Prob)$. Therefore, for $\imptheta = \impW-\FunOne$, we have,

\begin{align}\label{eq:eq_function}
    q_u:=\V_u(\Qrob) &= \int_{(\X\times\Y)} k_{u(x)}d\Qrob(x,y)=\E_\Qrob[k_{u}] \nonumber\\
    & \stackrel{a.s.}{=} E_\Qrob[E_\Qrob[k_{u}|\F_\Y]] \stackrel{a.s.}{=} E_\Qrob[E_\Prob[k_{u}|\F_\Y]\nonumber\\
    &\stackrel{a.s.}{=} E_\Prob[\impW E_\Prob[k_{u}|\F_\Y]] \stackrel{a.s.}{=} E_\Prob[\imptheta E_\Prob[k_{u}|\F_\Y]] + p_u \nonumber\\
    &\stackrel{a.s.}{=} E_\Prob[\kernel_Y^u\langle \kernel_Y,\imptheta\rangle]=\T_u\imptheta+p_u
\end{align}

where we assume $\imptheta\in\Hilbert$ for all $\measures$. Similar to the categorical setting, $\imptheta$ accounts for the amount of shift. 
We denote $\imptheta_{\max}$ as the maximum norm on allowed shifts. In the following we denote $\umax=\|\sup_{\mu\in\measures}\V_u(\mu)\|$, and consider the case where $\T_u$ has a bounded inverse.

\section{Estimation and Generalization}
In this section we provide a series of methods to estimate the effective shift weight $\imptheta$.

\subsection{Categorical Label Spaces}\label{subsec:categorical}
For categorical label spaces, we use Eq.~\ref{eq:eq_vector} to approximate $\imptheta\in\Re^k$. However, as mentioned before, we need to estimate the vectors $q_g,p_q\in\Re^d$ and matrix $T_g\in\Re^{d\times k}$ in $q_g-p_q= T_g\impthetaV $ equality. Let $\wh q_g,\wh p_g$, and $\wh T_g$ to be the estimates of $q_g, p_g$, and $ T_g$ respectively. We use $\Dqg$ to denote an upper bound 
(can be high probability upper bound) on $\|q_g-\wh q_q\|$,  $\Dpg$ on  $\|p_g-\wh p_q\|$, and $\DTg$ on $\|T_g-\wh T_q\|$. For $k\geq d$, let $\|T_g^\dagger\|$ denote the inverse of a smallest singular value of the matrix $T_g$, and $\imptheta_{\max}$ denote an upper bound on $\|\imptheta\|$ of the true $\imptheta$.

\begin{lemma}\label{lemma:inversV}
Consider a non-degenerate matrix $T_g$, vectors $q_g,p_g$, where $ \imptheta= T_g^\dagger ( q_g- p_g)$. Also consider the corresponding estimates, $\wh T_g,\wh q_g,\wh p_g$, and estimation errors $\DTg,\Dpg,\Dqg$. When  $\DTg\leq\frac{1}{2\|T_g^\dagger\|}$, for $\wh \imptheta=\wh T_g^\dagger (\wh q_g-\wh p_g)$,
\begin{align*}
    \|\wh \imptheta-\imptheta\|&\leq 2\|T_g^\dagger\| \left(\Dqg+\Dpg+\impthetamax\DTg\right)
\end{align*}
\end{lemma}
Proof \ref{proof:lemma:inversV}. We refer to this estimator $\wh \imptheta=\wh T_g^\dagger (\wh q_g-\wh p_g)$ as $\Eone$. Note that the estimate depends on the choice of $g$, and how well-conditioned the forward operator $\|T_g\|$ is, i.e., how small $\|T_g^\dagger\|$ is. For the sake of notation simplicity, we drop the dependence in $g$ (and later in $u$).

We obtain $\wh q_g,\wh p_q$, and $\wh T_g$  by applying function $g$ to $\alpha n$ data points in $\DS$, and $m$ data points in $\DT$. In other words,
\begin{align}\label{eq:FiniteEstimator}
    \!\!\!\wh q_g &= \E_{\wh\Qrob_m}[g(X)],~\wh p_g = \E_{\wh\Prob_{\alpha n}}[g(X)],~\wh T_g = \E_{\wh\Prob_{\alpha n}}[g(X)e_Y^\top]\!
\end{align}
where, $\forall i\in[k]$, $e_i\in\Re^k$ is the $i$'th standard basis vector, with all elements are zero except the $i$'th element is one. 
\begin{lemma}\label{lemma:FiniteConcentration}
Using the estimates in Eq.~\ref{eq:FiniteEstimator}, we have
\begin{align}
    \Dpg &\leq\sqrt{\frac{d}{\alpha n} \log(\frac{2d}{\delta})},~\Dqg \leq \sqrt{\frac{d}{m} \log(\frac{2d}{\delta})},~\DTg \leq 2\sqrt{\frac{2d}{\alpha n}\log(\frac{2(d+k)}{\delta})}
\end{align}
each with probability at least $1-\delta$.
\end{lemma}
Lemma~\ref{lemma:FiniteConcentration} follows from the standard application of Hoeffding's inequality to vectors, and concentration of adjoint matrices and dilation (Thm. 1.3 and section 2.6 in \citet{tropp2012user}).
\begin{theorem}\label{theorem:inverse}
Using the direct estimator $\wh \imptheta=\wh T_g^\dagger (\wh q_g-\wh p_g)$, when the number of samples $n\geq \frac{32}{\alpha}\|T_g^\dagger\|^2d\log(\frac{6(d+k)}{\delta})$,
\begin{align*}
    \|\wh \imptheta-\imptheta\|&\leq \epsilon(\delta):=2\|T_g^\dagger\| \Big(\sqrt{\frac{d}{\alpha n} \log(\frac{6d}{\delta})}+\sqrt{\frac{d}{m} \log(\frac{6d}{\delta})}+ 2\impthetamax\sqrt{\frac{2d}{\alpha n}\log(\frac{6(d+k)}{\delta})}\Big)
\end{align*}
with probability at least $1-\delta$.
\end{theorem}
The statement in the Theorem~\ref{theorem:inverse} directly follows from the statements in Lemma~\ref{lemma:inversV} and Lemma~\ref{lemma:FiniteConcentration}.

\begin{remark}[Comparison of Theorem~\ref{theorem:inverse} with \citet{lipton2018detecting}] \citet{lipton2018detecting} propose to use $\wh \impW=\wh T_g^\dagger \wh q_g$ to estimate the importance weight vector. When $n$ is large enough which is similar to the condition in the Theorem \ref{theorem:inverse}, the authors provide a concentration bound on $\|\wh \impW-\impW\|$ which holds under the following additional conditions: $(i)$ $T_g$ has to be square matrix, i.e., $d=k$. 
$(ii)$ the estimation error in $\DTg$ should be full rank which is an strong requirement, $(iii)$ $g$ needs to be a classifier. The proposed analysis in this paper does not require any of these assumptions, and furthermore, it improves the bound provided in~\citet{lipton2018detecting}.
\end{remark}

An alternative approach to estimate the  $\wh \imptheta$ is to use regularized approximation. Regularized approach has been proposed in \citet{azizzadenesheli2019regularized} for the special case when $g$ is a classifier. In this paper we generalize this approach to general functions. The underlying optimization for the regularized approach is as follows,
\begin{align}\label{eq:reg}
    \wh \imptheta = \argmin_{\imptheta'} \|\wh T_g\imptheta'-\wh q_g+\wh p_g\|+\DTg\|\imptheta'\|
\end{align}
We refer to this estimator as $\Etwo$.
\begin{lemma}\label{lemma:regularizedV}
Consider a non-degenerate matrix $T_g$, vectors $q_g,p_g$, where $ \imptheta= T_g^\dagger ( q_g- p_g)$. Also consider the corresponding estimates, $\wh T_g,\wh q_g,\wh p_g$, and estimation errors $\DTg,\Dpg,\Dqg$. For $\wh \imptheta$, a solution to Eq.~\ref{eq:reg}, we have,
\begin{align*}
    \|\wh \imptheta-\imptheta\|&\leq 2\|T_g^\dagger\| (\Dqg+\Dpg+\impthetamax\DTg)
\end{align*}
\end{lemma}
Proof~\ref{proof:lemma:regularizedV}. It is important to note that the right hand side of both bounds in Lemma~\ref{lemma:inversV} and Lemma~\ref{lemma:regularizedV} are equal and the only difference is on the burning time required in Lemma~\ref{lemma:inversV}. Such direct comparison is not possible using the bound provided in~\citep{lipton2018detecting}. The Lemma~\ref{lemma:inversV} and Lemma~\ref{lemma:regularizedV} indicate that regularized estimation method is more favorable for two main reasons; $(i)$ it does not require minimum number of samples, $(ii)$ the minimum number of samples required in direct inverse approach depends on a priori unknown parameters of the problem, which the regularization based approach does not require such prior knowledge. 

\begin{theorem}\label{theorem:RLLS}
The regularized estimator in Eq.~\ref{eq:reg} satisfies,
\begin{align*}
    \|\wh \imptheta-\imptheta\|&\leq\epsilon(\delta):= 2\|T_g^\dagger\| \Big(\sqrt{\frac{d}{\alpha n} \log(\frac{6d}{\delta})}+\sqrt{\frac{d}{m} \log(\frac{6d}{\delta})}+ 2\impthetamax\sqrt{\frac{2d}{\alpha n}\log(\frac{6(d+k)}{\delta})}\Big)
\end{align*}
with probability at least $1-\delta$.
\end{theorem}
This Theorem directly follows from Lemma~\ref{lemma:FiniteConcentration} and Lemma~\ref{lemma:regularizedV}. Similar to \citet{lipton2018detecting}, \citet{azizzadenesheli2019regularized} also uses $g\in\FuncClass$, and the results in the Theorem \ref{theorem:RLLS} is the generalization of the prior work to general functions $g$.

Note that the above mentioned bounds depends on $\|T_g^\dagger\|$. Prior works \citep{lipton2018detecting,azizzadenesheli2019regularized} realized Eq.~\ref{eq:finite_confusion} just for the special case of $g\in\FuncClass$, In this case, the square matrix $T_g$ has a special form of confusion matrix, i.e., $(T_g)_{i,j} = \Prob(g(X)=i,Y=j)$ and has entries sum to one. In the best case, where the $g$ is a perfect classifier with zero loss in the special case of realizable setting, $\|T_g^\dagger\|=k$, otherwise $\|T_g^\dagger\|\geq k$. However, in this paper, when we allow more general $g$ to exploit the spectrum more appropriately, e.g., orthogonal points on a sphere (or cube), then $\|T_g^\dagger\|=1$. Therefore, the bounds in the Theorems~\ref{theorem:inverse}, and Theorems~\ref{theorem:RLLS} further improve the prior bounds in \citet{lipton2018detecting,azizzadenesheli2019regularized}.

\subsection{General Normed Label Spaces}

For normed label space, we use Eq.~\ref{eq:eq_function} to approximate $\imptheta\in\Hilbert$. However, as mentioned before, we need to estimate the functions $q_u,p_u\in\Hilbert$ and the operator $\T_u\in\HS$ in the $q_u-p_u= T_g\imptheta$ equality. Let $\wh q_u,\wh p_u$, and $\wh \T_u$ be the estimates respectively. We use $\Dqu$ to denote an upper bound (e.g., high probability) on $\|q_u-\wh q_u\|$,  $\Dpu$ on  $\|p_u-\wh p_u\|$, and $\DTu$ on $\|\T_u-\wh \T_u\|$.
Let $\imptheta_{\max}$ denote an upper bound on $\|\imptheta\|$ of the true $\imptheta$. For small enough $\DTu$, when $\|\T_u^{-1}(\wh \T_u-\T_u)\|<1$, we have that $\wh \T_u^{-1}$ exist and,
\begin{align*}
    \|\wh \T_u^{-1}\|\leq \frac{\|\T_u^{-1}\|}{1-\|\T_u^{-1}(\wh \T_u-\T_u)\|}
\end{align*}
Therefore, under sufficiently small $\DTu$, the direct estimate of $\imptheta$ is the following estimator,
\begin{align}\label{eq:E3}
     \!\!\!\!\!\!\!\!\!\text{ $\Ethree$:}\quad\quad\quad\wh \imptheta=\wh \T_u^{-1}(\wh q_u-\wh p_u)\quad\quad\quad\quad
\end{align}
\begin{lemma}\label{lemma:inversVFunc}
Consider $ \imptheta= T_g^\dagger ( q_g- p_g)$. For the estimates, $\wh T_g,\wh q_g,\wh p_g$, and estimation errors $\DTg,\Dpg,\Dqg$, when $\DTu\leq \frac{1}{2\|\T_u^{-1}\|}$, then $\wh \imptheta$, the solution to Eq.~\ref{eq:eq_function}, satisfies,
\begin{align*}
    \|\wh \imptheta-\imptheta\|&\leq 2\|\T_u^{-1}\| (\Dqu+\Dpu+\DTu\imptheta_{\max})
\end{align*}
\end{lemma}
Proof~\ref{proof:lemma:inversVFunc}. We obtain $\wh q_u,\wh p_u$, and $\wh \T_u$  by applying function $u$ to $\alpha n$ data points in $\DS$, and $m$ data points in $\DT$. Ergo,
\begin{align}\label{eq:FuncEstimator}
    \wh q_u = \V_u(\wh\Qrob_{m}), \wh p_g = \V_u(\wh\Prob_{\alpha n}), \wh \T_u = \E_{\wh\Prob_{\alpha n}}[\kernel_{u(X)}\otimes k_Y],
\end{align}
Note that $\wh q_u$ and $\wh p_u$ are in the RKHS $\Hilbert$ therefore in a  Hilbert space with norm $\|\cdot\|$, and $\wh \T_u$ is in the Hilbert-Schmidt $\HS$, which it self is a Hilbert space with norm $\|\cdot\|_{\HS}$. In the following we deploy concentrations on Hilbert spaces. 
\begin{lemma}\label{lemma:FuncConcentration}
Using the estimates in Eq.~\ref{eq:FuncEstimator}, we have
\begin{align}
    \Dpu &\leq2\barkernel\sqrt{\frac{2}{\alpha n} \log(\frac{2}{\delta})},~
    \Dqg \leq 2\barkernel\sqrt{\frac{2}{m}\log(\frac{2}{\delta})},~\DTu \leq 2\barkernel\sqrt{\frac{2}{\alpha n}\log(\frac{2}{\delta})}
\end{align}
each with probability at least $1-\delta$.
\end{lemma}
Proof~\ref{proof:lemma:FuncConcentration}. The proof follows from the concentration inequalities in Hilbert space (in general Banach spaces)~\citep{pinelis1992approach,rosasco2010learning} and is provided in the appendix.

\begin{theorem}\label{theorem:inverseFunc}
Using the direct estimator $\wh \imptheta=\wh \T_u^{-1}(\wh q_u-\wh p_u)$, as the number of samples $n\geq \frac{32}{\alpha}\|\T_u^\dagger\|^2\barkernel^2\log(\frac{6}{\delta})$, then
\begin{align*}
    \|\wh \imptheta-\imptheta\|&\leq \epsilon(\delta):=4\|\T_u^\dagger\| \Big(\barkernel\sqrt{\frac{2}{\alpha n} \log(\frac{6}{\delta})}+ \barkernel\sqrt{\frac{2}{m}\log(\frac{6}{\delta})}+\impthetamax \barkernel\sqrt{\frac{2}{\alpha n}\log(\frac{6}{\delta})}\Big)
\end{align*}
with probability at least $1-\delta$.
\end{theorem}
The Theorem~\ref{theorem:inverseFunc} directly follows from the statements in Lemma~\ref{lemma:inversVFunc} and Lemma~\ref{lemma:FuncConcentration}. We propose an alternative approach to estimate the  $\wh \imptheta$. This approach is based on using regularized approximation. We propose the following regularized optimization problem, the estimator $\Efour$,
\begin{align}\label{eq:regFun}
    \wh \imptheta = \argmin_{\imptheta'} \left(\|\wh \T_u\imptheta'-\wh q_u+\wh p_u\|+\DTu\|\imptheta'\|\right)
\end{align}
This optimization is designed such that the outcome $\wh \imptheta$ minimizes the error in the desired objective $\|\wh \T_u\imptheta'-\wh q_u+\wh p_u\|$ while regularizing the shift to zero.

\begin{lemma}\label{lemma:regularizedVFunc}
Consider an operator $\T_u$, functions $q_u,p_u$, where $ \imptheta= \T_u^{-1} ( q_u- p_u)$. Also consider the corresponding estimates, $\wh \T_u,\wh q_u,\wh p_u$, and estimation errors $\DTu,\Dpu,\Dqu$. For $\wh \imptheta$, a solution to Eq.~\ref{eq:regFun}, we have:
\begin{align*}
    \|\T_u(\wh\imptheta-\imptheta)\|&\leq\left(\min_{\imptheta'}\| \T_u\imptheta'- q_u+ p_u\|+2\DTu\|\imptheta'\|\right) + 2(\Dpu + \Dqu)\\
\textit{and,}~
    \|\wh \imptheta-\imptheta\|&\leq 2\|\T_u^{-1}\| (\Dqu+\Dpu+\impthetamax\DTu)
\end{align*}
\end{lemma}
Proof~\ref{proof:lemma:regularizedVFunc}. The Lemma~\ref{lemma:regularizedVFunc} is that of independent importance and is the infinite dimension extension of finite sample value learning in the field of reinforcement learning~\citep{pires2012statistical}. Similar to categorical setting, it is important to note that the right hand side of both bounds in Lemma~\ref{lemma:inversVFunc} and Lemma~\ref{lemma:regularizedVFunc} are equal and the only difference is on the burning time required in Lemma~\ref{lemma:inversVFunc}.  The Lemma~\ref{lemma:inversVFunc} and Lemma~\ref{lemma:regularizedVFunc} indicate that regularized estimation method is more favorable for two main reasons; $(i)$ it does not have the minimum number of samples requirement, $(ii)$ the minimum number of samples required in direct inverse approach depends on a priori unknown parameters of the problem, which is not needed in the regularized approach of Eq.~\ref{eq:regFun}. 

\begin{theorem}\label{Theorem:RegFunc}
The regularized estimator in Eq.~\ref{eq:regFun} satisfies,
\begin{align*}
   \|\wh \imptheta-\imptheta\|&\leq \epsilon(\delta):= 4\|\T_u^\dagger\| \Big(\barkernel\sqrt{\frac{2}{\alpha n} \log(\frac{6}{\delta})} + \barkernel\sqrt{\frac{2}{m}\log(\frac{6}{\delta})}+\impthetamax \barkernel\sqrt{\frac{2}{\alpha n}\log(\frac{6}{\delta})}\Big)
\end{align*}
with probability at least $1-\delta$.
\end{theorem}
The statement of Theorem~\ref{Theorem:RegFunc} follows from statements in  Lemma~\ref{lemma:FuncConcentration} and Lemma~\ref{lemma:regularizedVFunc}. In the appendix \ref{apx:neural_operator} we provide a further discussion on how one can use deep neural networks in place of kernel $\kernel$ in estimating the importance weight.

\subsection{Generalization}
Having access to an estimate of importance weight, we deploy importance weighted \ERM to learn a predictor. As motivated by \citet{azizzadenesheli2019regularized}, when, for instance, the number samples from the target domain is small or the maximum expected shift is high, i.e., large $\impthetamax$, but at the same time, the number of required samples to have a reasonably small $\epsilon(\delta)$ is not much higher than the number of samples provided, the $\wh\imptheta$ estimate may not be a reliable estimate to be used. In this case, we may leave the $\wh\imptheta$ and stick to the best empirical risk minimizer on the source domain, i.e., we set $\wh\imptheta$ to zero ( $\wh\impW$ to ones) in Eq.\ref{eq:riskminimizer}. Motivated by such consideration, we use regularized importance weight in the empirical risk minimization, i.e., for $\gamma\geq 0$ we use, 
\begin{align}\label{eq:labelassignemnt}
\wh \impW_\gamma := 
  \begin{cases}
    \VecOne + \gamma \wh \imptheta, &\quad\quad\quad\quad\quad\textit{categorical label spaces} \\
    \FunOne + \gamma \wh \imptheta, &\quad\quad\quad\quad\quad\textit{normed label spaces.}~
    \end{cases}
\end{align}
For a function $\impW$, define a set of weighed loss functions,
\begin{align*}
\!\!\G(\ell,\FuncClass)\!:=\!\{l_f\!:\! l_f(x,y)\!=\!\impW(y)\ell(y,f(x)), \!\forall f\!\in\FuncClass,\forall (x,y)\!\in\!\X\!\times\!\Y\}    
\end{align*}
and its Rademacher complexity as follows,
\begin{align*}
    \Rade_{n}(\G)=\E_{(X_i,Y_i)\sim\Prob:i\in[n]}\left[\E_{\xi_i:i\in[ n]}\left[\frac{1}{ n}\sup_{f\in\FuncClass}\sum_i^{ n}\xi_il_f(X_i,Y_i)\right]\right]
\end{align*}
where $\{\xi_i\}_{i=1}^{n}$ is a collection of independent Rademacher random variables~\citep{bartlett2002rademacher}. Employing any of the estimators in Algorithm \ref{algo} to come up with a $\wh f$, and the statement of Theorem 1 in \citet{azizzadenesheli2019regularized}, we have,
\begin{theorem}[Generalization Guarantee]\label{theorem:gen}
For two probability measures $\Prob$ and $\Qrob$ on a measure space $(\X\times\Y,\F)$, consider $n$ sample from the source and $m$ sample from the target domain, the algorithm \ref{algo} outputs $\wh f\in\FuncClass$, for which we have,
\begin{align*}
     \loss(\wh f,\Qrob)-\inf_{f\in\FuncClass}\loss(f,\Qrob)\leq &\gamma\epsilon(\delta) + (1-\gamma)\impthetamax + 2 \Rade_{\alpha n}(\G)\\
     &\!\!\!\!\!\!\!\!\!\!\!\!\!\!\!\!\!\!\!\!\!\!\!\!\!\!\!\!\!\!\!\!\!\!\!\!\!\!\!\!\!\!\!\!\!\!\!\!\!\!\!\!\!\!\!+\min\Big\{ \dinf(\Qrob_\Y||\Prob_\Y)\sqrt{\frac{1}{\alpha n}\log(\frac{2}{\delta})}, \frac{\dinf(\Qrob_\Y||\Prob_\Y)}{\alpha n}\log(\frac{2}{\delta}) + \sqrt{\frac{2d(\Qrob_\Y||\Prob_\Y)}{\alpha n}\log(\frac{2}{\delta})} \Big\}
\end{align*}
with probability at least $1-3\delta$. If direct inverse method, $\Eone$ or $\Ethree$, is used, the above bound holds when $n$ satisfies $n\geq \frac{32}{\alpha}\|T_g^\dagger\|^2d\log(\frac{6(d+k)}{\delta})$ for categorical lalel spaces, and
$n\geq \frac{32}{\alpha}\|\T_u^\dagger\|^2\barkernel^2\log(\frac{6}{\delta})$ for normed label spaces.
\end{theorem}
The proof of Theorem \ref{theorem:gen} follows from MDFR Lemma (Lemma 4), and Theorem 1 in \citet{azizzadenesheli2019regularized}.


\begin{algorithm}[t!]
\caption{Importance Weighted ERM}
  \begin{algorithmic}[1]
\STATE \textbf{Inputs:} $\alpha$, $\gamma$, $\DS$, $\DT$, and $g$ or $u$,
\STATE Estimate $\wh\imptheta$ using $\alpha n$ and $m$ samples from $\DS$ and $\DT$,
\STATE Set $\wh\impW_\gamma$ according to Eq.~\ref{eq:labelassignemnt}
\STATE Return $\wh f\in\argmin_{f\in\FuncClass}\loss(f,\wh\Prob_n,\wh\impW_\gamma).$
  \end{algorithmic}
 \label{algo} 
\end{algorithm}

\section{Experiment}

We empirical study the performance of proposed importance weight estimators, in particular, $\Etwo$,~Eq.~\ref{eq:reg} for categorical, and $\Efour$,~Eq~\ref{eq:regFun} for normed spaces on synthetic data. 
\begin{figure}[h]
    \begin{subfigure}[t]{0.5\columnwidth}
    \centering
    \includegraphics[scale=0.4]{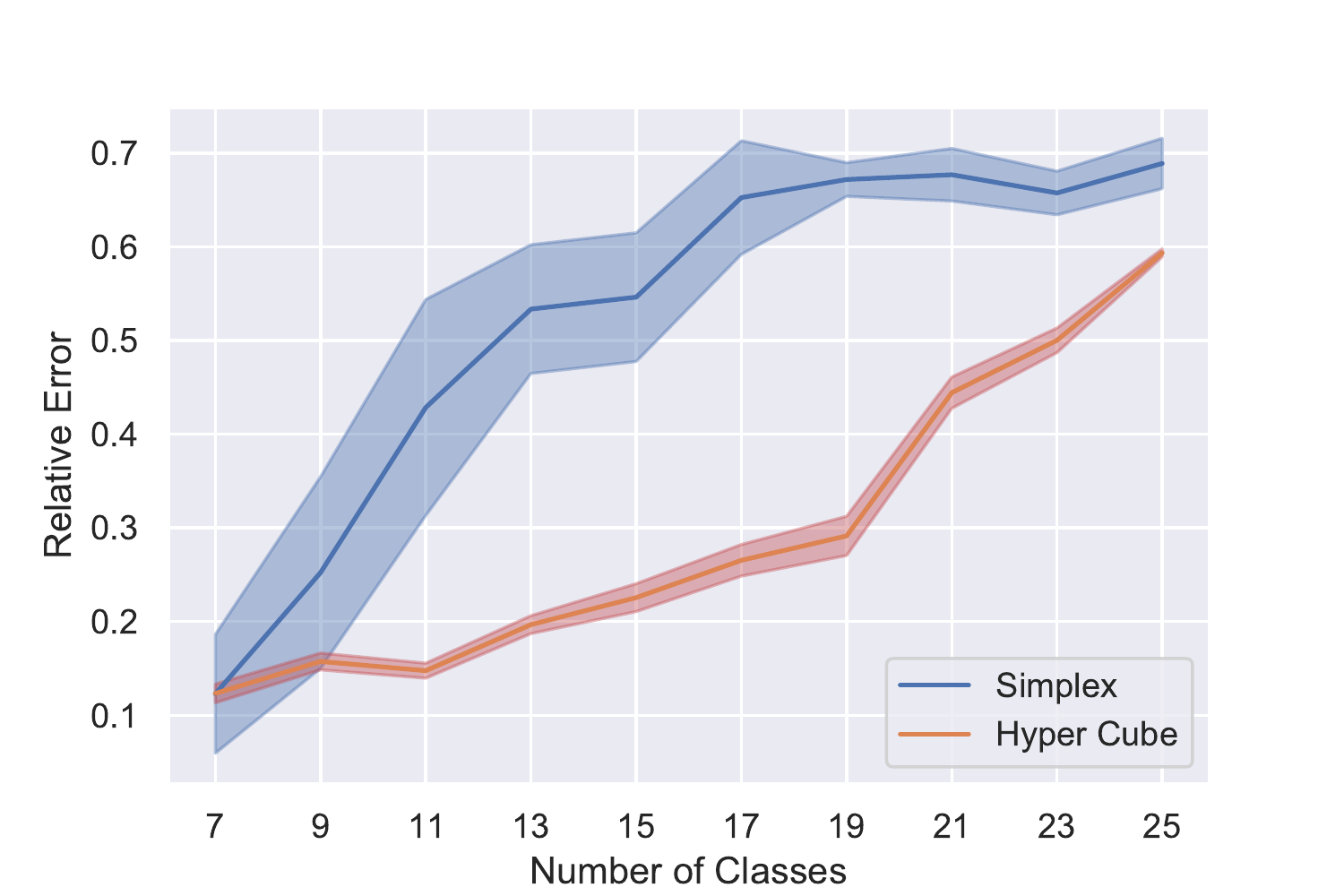}
    \end{subfigure}
    \begin{subfigure}[t]{0.5\columnwidth}
    \centering
    \includegraphics[scale=0.4]{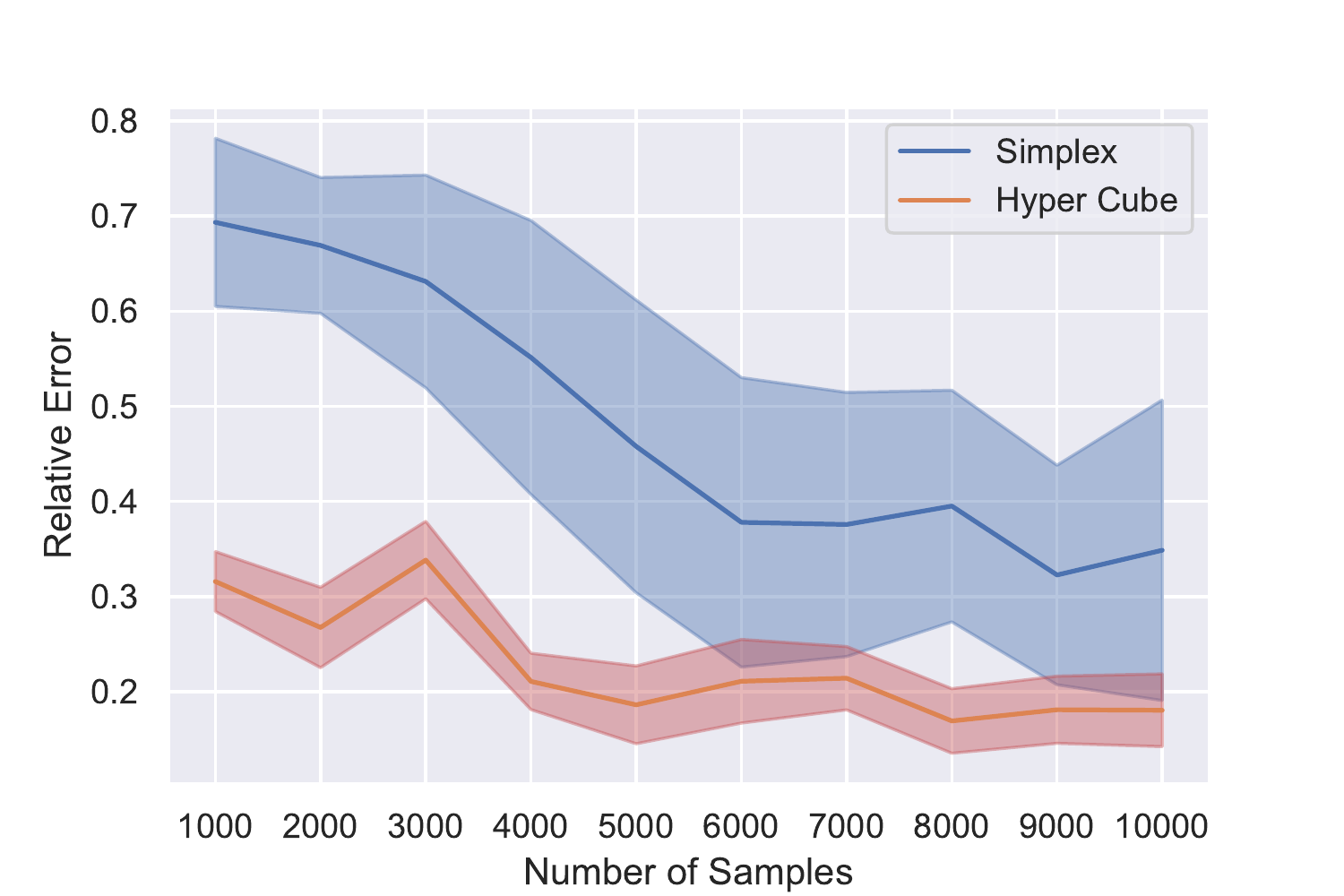}
    \end{subfigure}
    \vspace{-0.6cm}
    \caption{Categorical $\Y$}
    \label{categoriacal}
\end{figure}

Prior works provide an empirical study for estimator in Eq.~\ref{eq:reg} when the $g$ function is a classifier. As discussed in subsection~\ref{subsec:categorical}, and prescribed by the generalization in Theorem~\ref{theorem:RLLS}, allowing $g$ to be general form function might enhance the weight estimation. In the following, we provide comparisons in the estimation error of the estimator in Eq.~\ref{eq:reg}, $1)$ when $g$ is a deep neural network classifier with a softmax layer in the last layer with output on a \textbf{Simplex} and trained using cross entropy loss, and $2)$ $g$ is a same neural network with no softmax layer, and trained using one hot encoding of label, therefore, output as corners of a \textbf{Hyper Cube} using L2 loss. We first study the case where the number of data points is fixed, but the number of classes varies~Fig~\ref{categoriacal}(left) with y-axis as the relative estimation error of importance weights. 
As indicated in Fig~\ref{categoriacal}(left), using \textbf{Hyper Cube} provides a more consistent weight estimation compared with using \textbf{Simplex}. As the number of classes grows, we observe that both of these methods result in high error which is due to insufficient number of samples. In the second study, we keep the number of classes to be $20$, and increase the number of samples~Fig~\ref{categoriacal}(right) and observe \textbf{Hyper Cube} provides a much better samples complexity and recovers the importance weight with much fewer number of samples compared with \textbf{Simplex} .  
\begin{figure}[h]
    \begin{subfigure}[t]{0.5\columnwidth}
    \centering
    \includegraphics[scale=0.4]{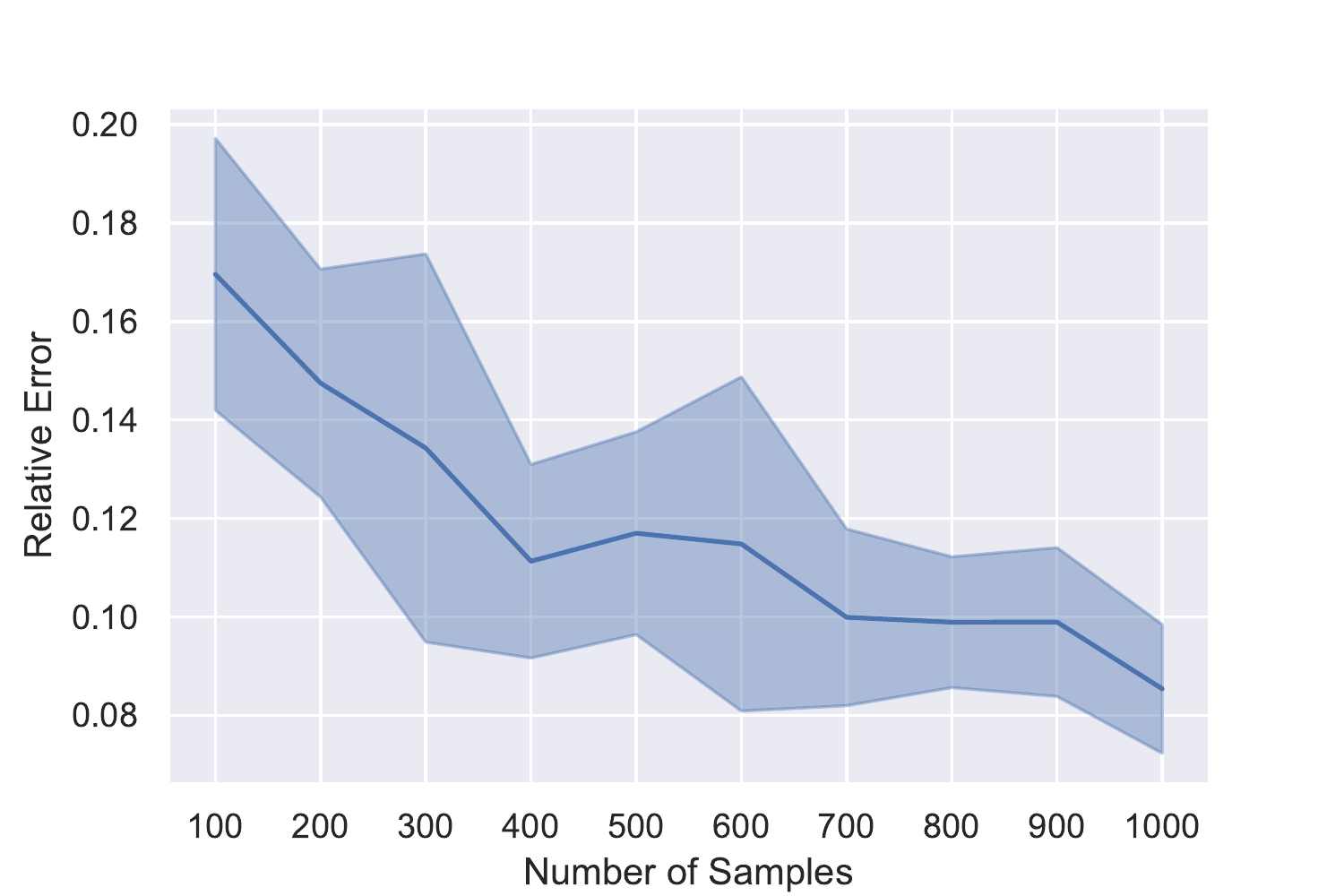}
    \end{subfigure}
    \begin{subfigure}[t]{0.5\columnwidth}
    \centering
\includegraphics[scale=0.4]{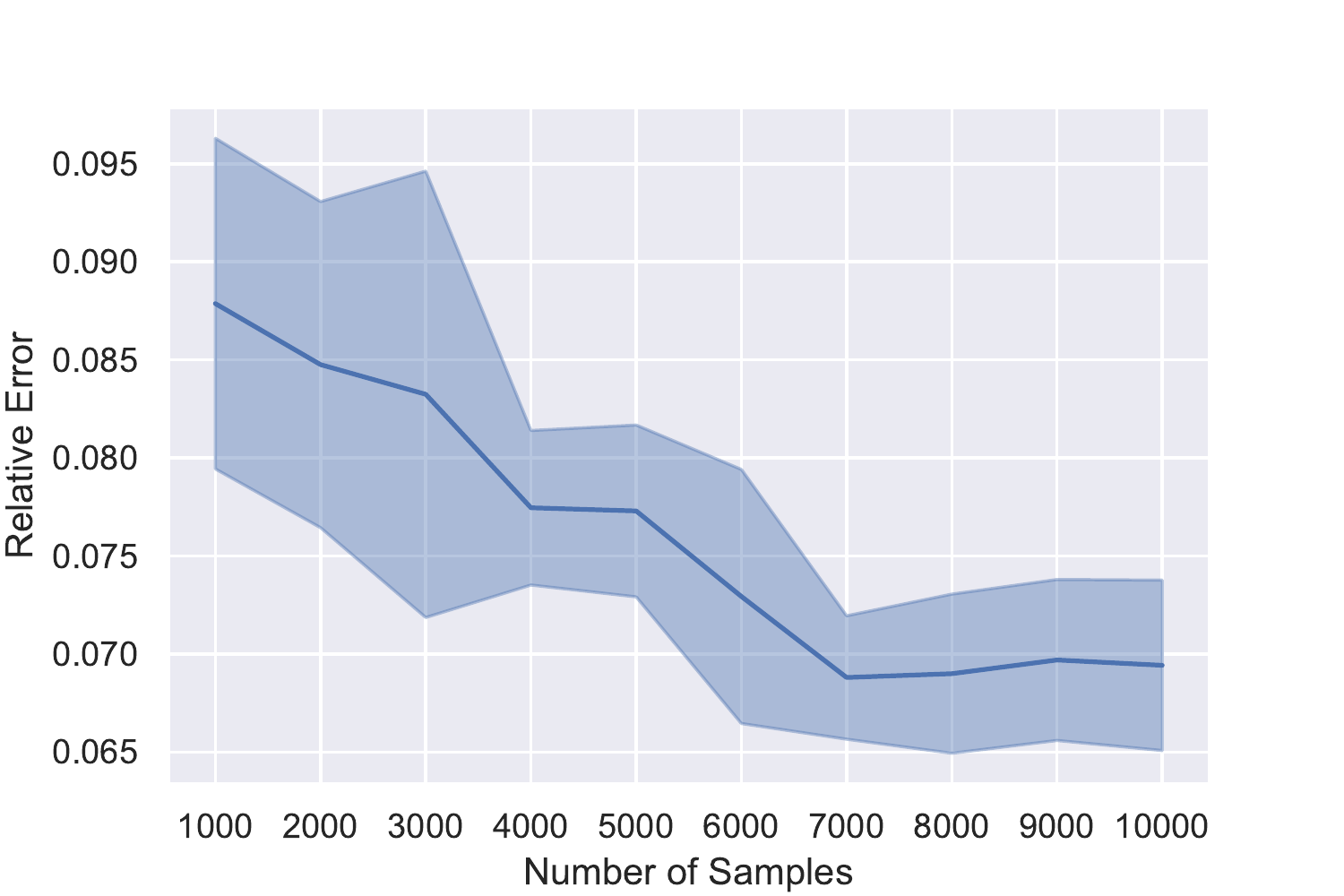}
    \end{subfigure}
    \vspace{-0.6cm}
    \caption{Normed vector space $\Y$}
    \label{fig:Normed}
\end{figure}
In Fig.~\ref{fig:Normed} we present the result when $\Y=\Re$. In this case, we use GP regression methods to learn the mapping from $\X$ to $\Y$, and compute the quantities in Eq.~\ref{eq:FuncEstimator}. We use squared version of Eq.~\ref{eq:regFun} to estimate $\impW$. The Fig.~\ref{fig:Normed} express that, as the number of samples increases, the estimation error improves. Finally, we made the code available for further use\footnote{\href{https://github.com/kazizzad/LabelShiftEstimator}{https://github.com/kazizzad/LabelShiftEstimator}} for further information. Please refer to appendix~\ref{Apx:exp} for details.

\section{Related Works}
\begin{sloppypar}
Domain adaptation is study of adapting to a new domain (target) under minimal access to labeled or unlabeled data from it~\citep{ben2010theory}. In standard supervised learning, the source and target follow the same measure ~\cite{vapnik1999overview,bartlett2002rademacher}. In the case of arbitrary shifts in the measures,~\citet{ben2010theory} introduces a notion of H-divergence to derive generalization analysis, where the multiple sources setting has been studied~\citep{crammer2008learning}. Robustness against distributional shifts has been widely studied in the literature which does not make explicit modeling assumptions on the shift~\cite{esfahani2017data, namkoong2016stochastic}). Moreover, under the covariate shift, adversarial approaches have been studied to developed robust model~\citep{liu2014robust,chen2016robust} where robustness is achieve only against very small changes in distributions to maintain sufficient performance.
\end{sloppypar}

\begin{sloppypar}
When the shift between two domain is not unstructured, the problem of covariate shift and label shifts have been considered. These settings become apparent in the context of casualty~\citep{scholkopf2012causal}. In this setting, $(i)$ the covariates causes the label (reward in contextual bandit), denoted as causal direction, and $(ii)$ the label can cause the symptoms (disease causes symptoms), denoted as anti-causal direction. 
When there is a shift in the measures, the knowledge of $\frac{d\Prob}{d\Qrob}$ can be deployed for importance weighted risk minimizing. 
In the setting of covariate shift, when $\frac{d\Prob_\X}{d\Qrob_\X}$ are known to the learner, the generalization power of importance weighted empirical risk minimization has been studied~\citep{zadrozny2004learning,cortes2010learning,cortes2014domain,shimodaira2000improving}. When $\frac{d\Prob_\X}{d\Qrob_\X}$ is not known, kernel methods have been deployed~\citep{huang2007correcting,gretton2009covariate,gretton2012kernel,zhang2013domain,zaremba2013b,shimodaira2000improving}. Such approaches fall short in high dimensional setting, especially images.
\end{sloppypar}

\begin{sloppypar}
Under some regularity condition, a measure over the covariates can be seen as a mixture of covariate conditional distribution. Prior works use this observation and under a strong assumption of pairwise mutual irreducibility~\citep{scott2013classification} show that, using Neyman-Pearson criterion~\citep{blanchard2010semi}, the mixture weights can be recovered for special cases~\cite{sanderson2014class,ramaswamy2016mixture,iyer2014maximum} which impose strong assumptions and computational challenges. In the presence of label shift, Bayesian methods are among others method that impose complicated computation requirements, e.g., \citep{storkey2009training, chan2005word} require computing posterior of label distribution for a given prior, treats the out come of a classifier as a probably distribution over labels, and
and suffers from lack of generalization guarantees.
\end{sloppypar}

\begin{sloppypar}
To address these challenges in label shift, also known as target shift~\citep{zhang2013domain}, and prior probability shift~\citep{moreno2012unifying,kull2014patterns,hofer2015adapting,tasche2017fisher},
a work by \citet{lipton2018detecting} propose black box correction method which is applicable to a wide range of label shift problems and drawn from  the classical grouping problem\cite{buck1966comparison,forman2008quantifying,saerens2002adjusting}. The authors provide a way to estimate the importance weights and use it for importance weighted empirical risk minimization. Following this idea, \citet{azizzadenesheli2019regularized} relaxes the assumptions required in \citep{lipton2018detecting} such as burning period, and full-rank assumption error matrix, and propose a regularized optimization method to estimate the importance weights up to a confidence interval. Using this confidence interval,~\citet{azizzadenesheli2019regularized} propose a novel analysis based on the second moment of importance weights and provides a first generalization guarantee for label shift. We utilize these development in our paper. 
\end{sloppypar}

\begin{sloppypar}
Under label shift setting, when the target domain has a balanced label distribution, but the source has imbalanced,~\citet{cao2019learning} proposes a margin based method to improve the generalization bound on the target domain. Calibration has been deployed for label shift problem~\citep{shrikumar2019calibration}, where it is later shown to be connected with importance weight approach~\citep{garg2020unified}. Recently,~\citet{kalan2020minimax} provides an study of transfer learning in the framework of deep neural networks, and~\citet{shui2020beyond} extend the H-divergence based guarantees to entropy based bounds.
\end{sloppypar}
\section{Conclusion}
In this paper, we study label shift and consider two cases of categorical and normed spaces of labels. We propose a suite of methods to estimate the importance weight from source to target domain only using unlabeled samples from the target and labeled samples from the source. We show that using such estimates results in desirable generalization properties. 

In our motivation examples, we discussed a medical setting where we use labeled data from a source country and send a group of volunteers to a target country with enough expertise and equipment to gather statistics of symptoms. Our task is to learn a good predictor for the target domain. Now consider a case that we have access to a limited number of specialists in the target country who can dedicate their time to diagnose patients. In case the diseases are not deadly, Which patients will we send to the doctors to be diagnosed with a goal to gain a good predictor with a few numbers of queries? In future work, we plan to study this setting where we need to actively decide whom to be diagnosed/investigated. This is an active learning domain adaptation under the label shift setting. 



\newpage
\bibliography{main}

\begin{thebibliography}{50}
\providecommand{\natexlab}[1]{#1}
\providecommand{\url}[1]{\texttt{#1}}
\expandafter\ifx\csname urlstyle\endcsname\relax
  \providecommand{\doi}[1]{doi: #1}\else
  \providecommand{\doi}{doi: \begingroup \urlstyle{rm}\Url}\fi

\bibitem[Azizzadenesheli et~al.(2019)Azizzadenesheli, Liu, Yang, and
  Anandkumar]{azizzadenesheli2019regularized}
Kamyar Azizzadenesheli, Anqi Liu, Fanny Yang, and Animashree Anandkumar.
\newblock Regularized learning for domain adaptation under label shifts.
\newblock \emph{arXiv preprint arXiv:1903.09734}, 2019.

\bibitem[Bartlett and Mendelson(2002)]{bartlett2002rademacher}
Peter~L Bartlett and Shahar Mendelson.
\newblock Rademacher and gaussian complexities: Risk bounds and structural
  results.
\newblock \emph{Journal of Machine Learning Research}, 2002.

\bibitem[Ben-David et~al.(2010)Ben-David, Blitzer, Crammer, Kulesza, Pereira,
  and Vaughan]{ben2010theory}
Shai Ben-David, John Blitzer, Koby Crammer, Alex Kulesza, Fernando Pereira, and
  Jennifer~Wortman Vaughan.
\newblock A theory of learning from different domains.
\newblock \emph{Machine learning}, 2010.

\bibitem[Blanchard et~al.(2010)Blanchard, Lee, and Scott]{blanchard2010semi}
Gilles Blanchard, Gyemin Lee, and Clayton Scott.
\newblock Semi-supervised novelty detection.
\newblock \emph{Journal of Machine Learning Research}, 11\penalty0
  (Nov):\penalty0 2973--3009, 2010.

\bibitem[Buck et~al.(1966)Buck, Gart, et~al.]{buck1966comparison}
AA~Buck, JJ~Gart, et~al.
\newblock Comparison of a screening test and a reference test in epidemiologic
  studies. ii. a probabilistic model for the comparison of diagnostic tests.
\newblock \emph{American Journal of Epidemiology}, 1966.

\bibitem[Cao et~al.(2019)Cao, Wei, Gaidon, Arechiga, and Ma]{cao2019learning}
Kaidi Cao, Colin Wei, Adrien Gaidon, Nikos Arechiga, and Tengyu Ma.
\newblock Learning imbalanced datasets with label-distribution-aware margin
  loss.
\newblock In \emph{Advances in Neural Information Processing Systems}, pages
  1567--1578, 2019.

\bibitem[Chan and Ng(2005)]{chan2005word}
Yee~Seng Chan and Hwee~Tou Ng.
\newblock Word sense disambiguation with distribution estimation.
\newblock In \emph{IJCAI}, 2005.

\bibitem[Chen et~al.(2016)Chen, Monfort, Liu, and Ziebart]{chen2016robust}
Xiangli Chen, Mathew Monfort, Anqi Liu, and Brian~D Ziebart.
\newblock Robust covariate shift regression.
\newblock In \emph{Artificial Intelligence and Statistics}, pages 1270--1279,
  2016.

\bibitem[Cortes and Mohri(2014)]{cortes2014domain}
Corinna Cortes and Mehryar Mohri.
\newblock Domain adaptation and sample bias correction theory and algorithm for
  regression.
\newblock \emph{Theoretical Computer Science}, 519:\penalty0 103--126, 2014.

\bibitem[Cortes et~al.(2010)Cortes, Mansour, and Mohri]{cortes2010learning}
Corinna Cortes, Yishay Mansour, and Mehryar Mohri.
\newblock Learning bounds for importance weighting.
\newblock In \emph{Advances in neural information processing systems}, pages
  442--450, 2010.

\bibitem[Crammer et~al.(2008)Crammer, Kearns, and Wortman]{crammer2008learning}
Koby Crammer, Michael Kearns, and Jennifer Wortman.
\newblock Learning from multiple sources.
\newblock \emph{Journal of Machine Learning Research}, 9\penalty0
  (Aug):\penalty0 1757--1774, 2008.

\bibitem[Cybenko(1989)]{cybenko1989approximation}
George Cybenko.
\newblock Approximation by superpositions of a sigmoidal function.
\newblock \emph{Mathematics of control, signals and systems}, 2\penalty0
  (4):\penalty0 303--314, 1989.

\bibitem[Esfahani and Kuhn(2018)]{esfahani2017data}
Peyman~Mohajerin Esfahani and Daniel Kuhn.
\newblock Data-driven distributionally robust optimization using the
  wasserstein metric: Performance guarantees and tractable reformulations.
\newblock \emph{Mathematical Programming}, 2018.

\bibitem[Forman(2008)]{forman2008quantifying}
George Forman.
\newblock Quantifying counts and costs via classification.
\newblock \emph{Data Mining and Knowledge Discovery}, 17\penalty0 (2):\penalty0
  164--206, 2008.

\bibitem[Garg et~al.(2020)Garg, Wu, Balakrishnan, and Lipton]{garg2020unified}
Saurabh Garg, Yifan Wu, Sivaraman Balakrishnan, and Zachary~C Lipton.
\newblock A unified view of label shift estimation.
\newblock \emph{arXiv preprint arXiv:2003.07554}, 2020.

\bibitem[Gretton et~al.(2009)Gretton, Smola, Huang, Schmittfull, Borgwardt, and
  Sch{\"o}lkopf]{gretton2009covariate}
Arthur Gretton, Alexander~J Smola, Jiayuan Huang, Marcel Schmittfull, Karsten~M
  Borgwardt, and Bernhard Sch{\"o}lkopf.
\newblock Covariate shift by kernel mean matching.
\newblock 2009.

\bibitem[Gretton et~al.(2012)Gretton, Borgwardt, Rasch, Sch{\"o}lkopf, and
  Smola]{gretton2012kernel}
Arthur Gretton, Karsten~M Borgwardt, Malte~J Rasch, Bernhard Sch{\"o}lkopf, and
  Alexander Smola.
\newblock A kernel two-sample test.
\newblock \emph{Journal of Machine Learning Research}, 13\penalty0 (Mar), 2012.

\bibitem[Hofer(2015)]{hofer2015adapting}
Vera Hofer.
\newblock Adapting a classification rule to local and global shift when only
  unlabelled data are available.
\newblock \emph{European Journal of Operational Research}, 243\penalty0
  (1):\penalty0 177--189, 2015.

\bibitem[Huang et~al.(2007)Huang, Gretton, Borgwardt, Sch{\"o}lkopf, and
  Smola]{huang2007correcting}
Jiayuan Huang, Arthur Gretton, Karsten~M Borgwardt, Bernhard Sch{\"o}lkopf, and
  Alex~J Smola.
\newblock Correcting sample selection bias by unlabeled data.
\newblock In \emph{Advances in neural information processing systems}, 2007.

\bibitem[Iyer et~al.(2014)Iyer, Nath, and Sarawagi]{iyer2014maximum}
Arun Iyer, Saketha Nath, and Sunita Sarawagi.
\newblock Maximum mean discrepancy for class ratio estimation: Convergence
  bounds and kernel selection.
\newblock In \emph{International Conference on Machine Learning}, pages
  530--538, 2014.

\bibitem[Kalan et~al.(2020)Kalan, Fabian, Avestimehr, and
  Soltanolkotabi]{kalan2020minimax}
Seyed Mohammadreza~Mousavi Kalan, Zalan Fabian, A~Salman Avestimehr, and Mahdi
  Soltanolkotabi.
\newblock Minimax lower bounds for transfer learning with linear and one-hidden
  layer neural networks.
\newblock \emph{arXiv preprint arXiv:2006.10581}, 2020.

\bibitem[Kress et~al.(1989)Kress, Maz'ya, and Kozlov]{kress1989linear}
Rainer Kress, V~Maz'ya, and V~Kozlov.
\newblock \emph{Linear integral equations}, volume~82.
\newblock Springer, 1989.

\bibitem[Kull and Flach(2014)]{kull2014patterns}
Meelis Kull and Peter Flach.
\newblock Patterns of dataset shift.
\newblock In \emph{First International Workshop on Learning over Multiple
  Contexts (LMCE) at ECML-PKDD}, 2014.

\bibitem[Lang(2012)]{lang2012real}
Serge Lang.
\newblock \emph{Real and functional analysis}, volume 142.
\newblock Springer Science \& Business Media, 2012.

\bibitem[Li et~al.(2020{\natexlab{a}})Li, Kovachki, Azizzadenesheli, Liu,
  Bhattacharya, Stuart, and Anandkumar]{li2020fourier}
Zongyi Li, Nikola Kovachki, Kamyar Azizzadenesheli, Burigede Liu, Kaushik
  Bhattacharya, Andrew Stuart, and Anima Anandkumar.
\newblock Fourier neural operator for parametric partial differential
  equations.
\newblock \emph{arXiv preprint arXiv:2010.08895}, 2020{\natexlab{a}}.

\bibitem[Li et~al.(2020{\natexlab{b}})Li, Kovachki, Azizzadenesheli, Liu,
  Bhattacharya, Stuart, and Anandkumar]{li2020neural}
Zongyi Li, Nikola Kovachki, Kamyar Azizzadenesheli, Burigede Liu, Kaushik
  Bhattacharya, Andrew Stuart, and Anima Anandkumar.
\newblock Neural operator: Graph kernel network for partial differential
  equations.
\newblock \emph{arXiv preprint arXiv:2003.03485}, 2020{\natexlab{b}}.

\bibitem[Li et~al.(2020{\natexlab{c}})Li, Kovachki, Azizzadenesheli, Liu,
  Stuart, Bhattacharya, and Anandkumar]{li2020multipole}
Zongyi Li, Nikola Kovachki, Kamyar Azizzadenesheli, Burigede Liu, Andrew
  Stuart, Kaushik Bhattacharya, and Anima Anandkumar.
\newblock Multipole graph neural operator for parametric partial differential
  equations.
\newblock \emph{Advances in Neural Information Processing Systems}, 33,
  2020{\natexlab{c}}.

\bibitem[Lipton et~al.(2018)Lipton, Wang, and Smola]{lipton2018detecting}
Zachary~C Lipton, Yu-Xiang Wang, and Alex Smola.
\newblock Detecting and correcting for label shift with black box predictors.
\newblock \emph{arXiv preprint arXiv:1802.03916}, 2018.

\bibitem[Liu and Ziebart(2014)]{liu2014robust}
Anqi Liu and Brian Ziebart.
\newblock Robust classification under sample selection bias.
\newblock In \emph{Advances in neural information processing systems}, pages
  37--45, 2014.

\bibitem[Moreno-Torres et~al.(2012)Moreno-Torres, Raeder, Alaiz-Rodr{\'\i}guez,
  Chawla, and Herrera]{moreno2012unifying}
Jose~G Moreno-Torres, Troy Raeder, Roc{\'\i}o Alaiz-Rodr{\'\i}guez, Nitesh~V
  Chawla, and Francisco Herrera.
\newblock A unifying view on dataset shift in classification.
\newblock \emph{Pattern recognition}, 2012.

\bibitem[Namkoong and Duchi(2016)]{namkoong2016stochastic}
Hongseok Namkoong and John~C Duchi.
\newblock Stochastic gradient methods for distributionally robust optimization
  with f-divergences.
\newblock In \emph{Advances in Neural Information Processing Systems}, pages
  2208--2216, 2016.

\bibitem[Nystr{\"o}m(1930)]{nystrom1930praktische}
Evert~J Nystr{\"o}m.
\newblock {\"U}ber die praktische aufl{\"o}sung von integralgleichungen mit
  anwendungen auf randwertaufgaben.
\newblock \emph{Acta Mathematica}, 1930.

\bibitem[Pinelis(1992)]{pinelis1992approach}
Iosif Pinelis.
\newblock An approach to inequalities for the distributions of
  infinite-dimensional martingales.
\newblock In \emph{Probability in Banach Spaces, 8: Proceedings of the Eighth
  International Conference}, pages 128--134. Springer, 1992.

\bibitem[Pires and Szepesv{\'a}ri(2012)]{pires2012statistical}
Bernardo~Avila Pires and Csaba Szepesv{\'a}ri.
\newblock Statistical linear estimation with penalized estimators: an
  application to reinforcement learning.
\newblock \emph{arXiv preprint arXiv:1206.6444}, 2012.

\bibitem[Ramaswamy et~al.(2016)Ramaswamy, Scott, and
  Tewari]{ramaswamy2016mixture}
Harish Ramaswamy, Clayton Scott, and Ambuj Tewari.
\newblock Mixture proportion estimation via kernel embeddings of distributions.
\newblock In \emph{International Conference on Machine Learning}, pages
  2052--2060, 2016.

\bibitem[Rosasco et~al.(2010)Rosasco, Belkin, and De~Vito]{rosasco2010learning}
Lorenzo Rosasco, Mikhail Belkin, and Ernesto De~Vito.
\newblock On learning with integral operators.
\newblock \emph{Journal of Machine Learning Research}, 11\penalty0 (2), 2010.

\bibitem[Saerens et~al.(2002)Saerens, Latinne, and
  Decaestecker]{saerens2002adjusting}
Marco Saerens, Patrice Latinne, and Christine Decaestecker.
\newblock Adjusting the outputs of a classifier to new a priori probabilities:
  a simple procedure.
\newblock \emph{Neural computation}, 14\penalty0 (1):\penalty0 21--41, 2002.

\bibitem[Sanderson and Scott(2014)]{sanderson2014class}
Tyler Sanderson and Clayton Scott.
\newblock Class proportion estimation with application to multiclass anomaly
  rejection.
\newblock In \emph{Artificial Intelligence and Statistics}, pages 850--858,
  2014.

\bibitem[Sch{\"o}lkopf et~al.(2012)Sch{\"o}lkopf, Janzing, Peters, Sgouritsa,
  Zhang, and Mooij]{scholkopf2012causal}
Bernhard Sch{\"o}lkopf, Dominik Janzing, Jonas Peters, Eleni Sgouritsa, Kun
  Zhang, and Joris Mooij.
\newblock On causal and anticausal learning.
\newblock \emph{arXiv preprint arXiv:1206.6471}, 2012.

\bibitem[Scott et~al.(2013)Scott, Blanchard, and
  Handy]{scott2013classification}
Clayton Scott, Gilles Blanchard, and Gregory Handy.
\newblock Classification with asymmetric label noise: Consistency and maximal
  denoising.
\newblock In \emph{Conference On Learning Theory}, 2013.

\bibitem[Shimodaira(2000)]{shimodaira2000improving}
Hidetoshi Shimodaira.
\newblock Improving predictive inference under covariate shift by weighting the
  log-likelihood function.
\newblock \emph{Journal of statistical planning and inference}, 2000.

\bibitem[Shrikumar and Kundaje(2019)]{shrikumar2019calibration}
Avanti Shrikumar and Anshul Kundaje.
\newblock Calibration with bias-corrected temperature scaling improves domain
  adaptation under label shift in modern neural networks.
\newblock \emph{arXiv preprint arXiv:1901.06852}, 1, 2019.

\bibitem[Shui et~al.(2020)Shui, Chen, Wen, Zhou, Gagn{\'e}, and
  Wang]{shui2020beyond}
Changjian Shui, Qi~Chen, Jun Wen, Fan Zhou, Christian Gagn{\'e}, and Boyu Wang.
\newblock Beyond $\mathcal{H}$-divergence: Domain adaptation theory with
  jensen-shannon divergence.
\newblock \emph{arXiv preprint arXiv:2007.15567}, 2020.

\bibitem[Storkey(2009)]{storkey2009training}
Amos Storkey.
\newblock When training and test sets are different: characterizing learning
  transfer.
\newblock \emph{Dataset shift in machine learning}, pages 3--28, 2009.

\bibitem[Tasche(2017)]{tasche2017fisher}
Dirk Tasche.
\newblock Fisher consistency for prior probability shift.
\newblock \emph{The Journal of Machine Learning Research}, 18\penalty0
  (1):\penalty0 3338--3369, 2017.

\bibitem[Tropp(2012)]{tropp2012user}
Joel~A Tropp.
\newblock User-friendly tail bounds for sums of random matrices.
\newblock \emph{Foundations of computational mathematics}, 12\penalty0
  (4):\penalty0 389--434, 2012.

\bibitem[Vapnik(1999)]{vapnik1999overview}
Vladimir~Naumovich Vapnik.
\newblock An overview of statistical learning theory.
\newblock \emph{IEEE transactions on neural networks}, 1999.

\bibitem[Zadrozny(2004)]{zadrozny2004learning}
Bianca Zadrozny.
\newblock Learning and evaluating classifiers under sample selection bias.
\newblock In \emph{Proceedings of the twenty-first international conference on
  Machine learning}. ACM, 2004.

\bibitem[Zaremba et~al.(2013)Zaremba, Gretton, and Blaschko]{zaremba2013b}
Wojciech Zaremba, Arthur Gretton, and Matthew Blaschko.
\newblock B-test: A non-parametric, low variance kernel two-sample test.
\newblock In \emph{Advances in neural information processing systems}, 2013.

\bibitem[Zhang et~al.(2013)Zhang, Sch{\"o}lkopf, Muandet, and
  Wang]{zhang2013domain}
Kun Zhang, Bernhard Sch{\"o}lkopf, Krikamol Muandet, and Zhikun Wang.
\newblock Domain adaptation under target and conditional shift.
\newblock In \emph{International Conference on Machine Learning}, 2013.

\end{thebibliography}
\bibliographystyle{plainnat}

\newpage
\appendix

\begin{center}
{\huge Appendix}
\end{center}




\section{Proofs}
\subsection{Proof of Lemma \ref{lemma:inversV}}\label{proof:lemma:inversV}
\begin{proof}[Lemma \ref{lemma:inversV}]
By definition, we have 
\begin{align*}
 \imptheta= T_g^\dagger ( q_g+ p_g),
\end{align*}
and  
\begin{align*}
\wh\imptheta= \wh T_g^\dagger ( \wh q_g+\wh p_g).    
\end{align*}
Therefore we have;
\begin{align}
\wh T_g(\wh \imptheta-\imptheta)&=\wh T_g\wh \imptheta- T_g \imptheta+ T_g \imptheta-\wh T_g \imptheta\nonumber\\
&=(\wh q_g- \wh p_g-(q_g- p_g))+ (T_g-\wh T_g) \imptheta.
\end{align}
Using Cauchy–Schwarz inequality, we have:
\begin{align}
\|\wh T_g(\wh \imptheta-\imptheta)\|&\leq\Dqg+\Dpg + \|(\wh T_g- T_g)\theta\|.
\end{align}
Using this statement, we have;
\begin{align}
\|\wh \imptheta-\imptheta\|&\leq 2\|T_g^\dagger\| \left(\Dqg+\Dpg+ \DTg\impthetamax\right).
\end{align}

\end{proof}

\subsection{Proof of Lemma \ref{lemma:regularizedV}}\label{proof:lemma:regularizedV}

\begin{proof}[Lemma \ref{lemma:regularizedV}]
Following the theorem 3.4 in \citet{pires2012statistical}, we have that, $\wh\imptheta$, the solution to minimization in Eq.~\ref{eq:reg}, satisfies, 
\begin{align}\label{eq:1}
    \|T_g\wh\imptheta-q_g+p_g\|\leq& \inf_{\imptheta'}\left(\|T_g\imptheta'-q_g+p_g\|+2\DTg\|\imptheta'\|\right)\nonumber\\
    &\quad\quad\quad+2(\Dpg+\Dqg).
\end{align}
The right hand side of Eq.~\ref{eq:1} is upper bounded by plugging in true $\theta$ instead of approaching the infimum, i.e.,
\begin{align}\label{eq:2}
    \|T_g\wh\imptheta-q_g+p_g\|&\leq \|T_g\imptheta-q_g+p_g\|+2\DTg\|\imptheta'\|\nonumber\\
    &\quad\quad\quad\quad\quad\quad\quad+2(\Dpg+\Dqg)\nonumber\\
    &=2\DTg\|\imptheta\|+2(\Dpg+\Dqg),
\end{align}
the last line follows since $T_g\imptheta=q_g-p_g$. Using the equality $T_g\imptheta=q_g-p_g$ one more time on the left hand side of Eq.~\ref{eq:2}, we have

\begin{align}
    \|T_g\wh\imptheta-q_g+p_g\|&=\|T_g(\wh\imptheta-\imptheta+\imptheta)-q_g+p_g\|\nonumber\\
    &=\|T_g(\wh\imptheta-\imptheta)+T_g\imptheta-q_g+p_g\|\nonumber\\
    &=\|T_g(\wh\imptheta-\imptheta)\|.
\end{align}
Therefore, 
\begin{align}
    \|T_g(\wh\imptheta-\imptheta)\leq 2\DTg\|\imptheta\|+2(\Dpg+\Dqg).
\end{align}
Resulting in
\begin{align}
    \|(\wh\imptheta-\imptheta)\leq 2\|T_g^\dagger\| \left(\DTg\impthetamax+\Dpg+\Dqg\right).
\end{align}
which is the statement of the Lemma~\ref{lemma:regularizedV}.

\end{proof}

\subsection{[Proof of Lemma \ref{lemma:inversVFunc}]}\label{proof:lemma:inversVFunc}
\begin{proof}[Lemma \ref{lemma:inversVFunc}]
By definition, since $\T_u$ has bounded inverse, we have 
\begin{align*}
\imptheta= \T_u^\dagger ( q_u+ p_u),    
\end{align*}
and since 
\begin{align*}
    \|\T_u^{-1}(\wh \T_u-\T_u)\|<1,
\end{align*}
we have
\begin{align*}
    \wh\imptheta= \wh \T_u^\dagger ( \wh q_u+\wh p_u).
\end{align*}
Therefore we have;
\begin{align}
\wh \T_u(\wh \imptheta-\imptheta)&=\wh \T_u\wh \imptheta- \T_u \imptheta+ \T_u \imptheta-\wh \T_u \imptheta\nonumber\\
&=(\wh q_u- \wh p_u-(q_u- p_u))+ (T_u-\wh T_u) \imptheta.
\end{align}
Using Cauchy–Schwarz inequality, we have:
\begin{align}
\|\wh \T_u(\wh \imptheta-\imptheta)\|&\leq\Dqu+\Dpu + \|(\wh T_u- T_u)\theta\|.
\end{align}
Using this statement, we have;
\begin{align}
\|\wh \imptheta-\imptheta\|&\leq 2\|\T_u^\dagger\| \left(\Dqu+\Dpu+ \DTu\impthetamax\right).
\end{align}

\end{proof}

\subsection{Proof of Lemma~\ref{lemma:FuncConcentration}}\label{proof:lemma:FuncConcentration}
\begin{proof}[Lemma \ref{lemma:FuncConcentration}]
For $\{\chi\}_{i}^t$, a collection of mean-zero independent random variables in a measure space of $\Hilbert$, if $\|\chi_i\|\leq c$, a.s., then,
\begin{align}
    \|\frac{1}{t}\sum_i^t \chi_i\|\leq c\sqrt{\frac{2}{n}\log(\frac{2}{\delta})},
\end{align}
with probability at least $1-\delta$~\citep{pinelis1992approach,rosasco2010learning}.

To develop the confidence interval for $\Dpu$ in $\|\wh p_u-p_u\|$ we set $t=\alpha n$, and $\chi_i = \kernel_{u(x_i)}-p_u$ for the $i$'th sample in the $\alpha n$ portion of data points in source data set $\DS$. Note that, $\chi_i = \kernel_{u(x_i)}-p_u$ is a mean-zero random variable for all $i$ and $\|\kernel_{u(X)}-p_u\|\leq 2\barkernel$. Ergo, $\E_{\wh\Prob_{\alpha n}}[\chi] =\E_{\wh\Prob_{\alpha n}}[\kernel_{u(X)]}-p_u$ and $c=2\barkernel$, resulting in the following 
\begin{align}
    \|\E_{\wh\Prob_{\alpha n}}[\kernel_{u(X)}]-p_u\|\leq 2\barkernel\sqrt{\frac{2}{\alpha n} \log(\frac{2}{\delta})},
\end{align}
with probability at least $1-\delta$. With a similar argument for $m$ data-points in the target data set $\DT$, we have 
\begin{align}
    \|\E_{\wh\Qrob_{m}}[\kernel_{u(X)}]-q_u\|\leq 2\barkernel\sqrt{\frac{2}{m} \log(\frac{2}{\delta})},
\end{align}
with probability at least $1-\delta$, and

\begin{align*}
    \|\E_{\wh\Prob_{\alpha n}}[\kernel_{u(X)}\otimes k_Y]
    - \T_u\|\leq 2\barkernel\sqrt{\frac{2}{\alpha n} \log(\frac{2}{\delta})},
\end{align*}
with probability at least $1-\delta$.
\end{proof}

\subsection{Proof of Lemma~\ref{lemma:regularizedVFunc}}\label{proof:lemma:regularizedVFunc}
\begin{lemma}\label{lemma:difference}
For any function $\imptheta'\in\Hilbert$, we have:
\begin{align}
    &\left|\|\T_u\imptheta'-q_u+p_u\|-\|\wh\T_u\imptheta'-\wh q_u+\wh p_u\|\right|\nonumber\\
    &\quad\quad\quad\quad\quad\quad\quad\quad\quad\quad\leq \DTu\|\imptheta'\| + \Dpu + \Dqu.
\end{align}
\end{lemma}
\begin{proof}[Lemma \ref{lemma:difference}]
Using triangle in equality twice we have, 
\begin{align}
    \|\T_u\imptheta'&-q_u+p_u\|=\|\T_u\imptheta'-q_u+p_u-(\wh\T_u\imptheta'-\wh q_u+\wh p_u)\nonumber\\
    &\quad\quad\quad\quad\quad\quad\quad\quad\quad+\wh\T_u\imptheta'-\wh q_u+\wh p_u\|\nonumber\\
    &\leq\|(\T_u-\wh\T_u)\imptheta'-(q_u-\wh q_u)+(p_u-\wh p_u)\|\nonumber\\
    &\quad\quad\quad\quad\quad\quad\quad\quad\quad+\|\wh\T_u\imptheta'-\wh q_u+\wh p_u\|,
\end{align}
therefore,
\begin{align}
    \|\T_u\imptheta'&-q_u+p_u\|-\|\wh\T_u\imptheta'-\wh q_u+\wh p_u\| \nonumber\\
    &\leq\|(\T_u-\wh\T_u)\imptheta'-(q_u-\wh q_u)+(p_u-\wh p_u)\|\nonumber\\
    &\leq\DTu\imptheta'+\Dqu+\Dpu.
\end{align}
With a similar argument for $\|\wh \T_u\imptheta'-\wh q_u+\wh p_u\|$ Similarly, we have,
\begin{align}
    \|\wh\T_u\imptheta'&-\wh q_u+\wh p_u\|=\|\wh \T_u\imptheta'-\wh q_u+\wh p_u-(\T_u\imptheta'- q_u+ p_u)\nonumber\\
    &\quad\quad\quad\quad\quad\quad\quad\quad\quad+\T_u\imptheta'- q_u+ p_u\|\nonumber\\
    &\leq\|(\wh \T_u-\T_u)\imptheta'-(\wh q_u- q_u)+(\wh p_u- p_u)\|\nonumber\\
    &\quad\quad\quad\quad\quad\quad\quad\quad\quad+\|\T_u\imptheta'- q_u+\wh p_u\|,
\end{align}
therefore,
\begin{align}
    \|\wh \T_u\imptheta'&-\wh q_u+\wh p_u\|-\|\T_u\imptheta'- q_u+ p_u\|\nonumber\\ 
    &\leq\|(\wh \T_u-\T_u)\imptheta'-(\wh q_u- q_u)+(\wh p_u- p_u)\|\nonumber\\
    &\leq\DTu\imptheta'+\Dqu+\Dpu.
\end{align}
Putting these inequalities together we have;
\begin{align}
    &\left|\|\wh \T_u\imptheta'-\wh q_u+\wh p_u\|-\|\T_u\imptheta'- q_u+ p_u\| \right|\nonumber\\
    &\quad\quad\quad\quad\quad\quad\quad\quad\quad\quad\leq\DTu\imptheta'+\Dqu+\Dpu,
\end{align}
which states the Lemma.
\end{proof}
Now we consider the following optimization. For a given $\lambda>0$, we define $\wh\imptheta_\lambda$ as follows;

\begin{align}\label{eq:lambda}
    \wh \imptheta_\lambda = \argmin_{\imptheta'} \|\wh \T_u\imptheta'-\wh q_u+\wh p_u\|+\lambda\|\imptheta'\|.
\end{align}

For $\wh\imptheta_\lambda$, the solution to Eq.~\ref{eq:lambda}, we have,
\begin{align}\label{eq:lambdaupperbund}
\|\wh \T_u \wh\imptheta_\lambda-\wh q_u+\wh p_u\|   \leq \min_{\imptheta'}\|\wh \T_u\imptheta'-\wh q_u+\wh p_u\|+\lambda\|\imptheta'\|,
\end{align}
and,
\begin{align}
\|\wh\imptheta_\lambda\| = \frac{1}{\lambda} \min_{\imptheta'}\|\wh \T_u\imptheta'&-\wh q_u+\wh p_u\|+\lambda\|\imptheta'\|\nonumber\\
&-\frac{1}{\lambda}\|\wh \T_u \wh\imptheta_\lambda-\wh q_u+\wh p_u\|.
\end{align}

\begin{lemma}\label{lemma:lambda}
For $\wh\imptheta_\lambda$, the solution to Eq.~\ref{eq:lambda}, we have,
\begin{align}
    &\|\T_u\wh\imptheta_\lambda-q_u+p_u\|\nonumber\\
    &\quad\quad\quad\quad\leq\max\{1,\frac{\DTu}{\lambda}\}\min_{\theta'}\big(\|\T_u\wh\imptheta_\lambda-q_u+p_u\|\nonumber\\
    &\quad\quad\quad\quad\quad\quad\quad\quad\quad\quad\quad\quad\quad\quad+(\DTu+\lambda)\|\imptheta'\|\big)\nonumber\\
    &\quad\quad\quad\quad+\max\{2,(1+\frac{\DTu}{\lambda})\}(\Dqu+\Dpu).
\end{align}
\end{lemma}
\begin{proof}[Lemma~\ref{lemma:lambda}]
Using the statement in the Lemma~\ref{lemma:difference} for $\wh\imptheta_\lambda$, we have
\begin{align}
    \|\T_u\wh\imptheta_\lambda&-q_u+p_u\|\nonumber\\
    &\leq \|\wh\T_u\wh\imptheta_\lambda-\wh q_u+\wh p_u\| + \DTu\|\wh\imptheta_\lambda\| + \Dpu + \Dqu\nonumber\\
    &=\|\wh\T_u\wh\imptheta_\lambda-\wh q_u+\wh p_u\|\nonumber\\
    &\quad\quad\quad\quad -\frac{\DTu}{\lambda}\|\wh\T_u\wh\imptheta_\lambda-\wh q_u+\wh p_u\|\nonumber\\
    &\quad\quad\quad\quad\quad\quad\quad\quad+\frac{\DTu}{\lambda}\|\wh\T_u\wh\imptheta_\lambda-\wh q_u+\wh p_u\| \nonumber\\
    &\quad\quad\quad\quad\quad\quad\quad\quad\quad\quad\quad\quad+ \DTu\|\wh\imptheta_\lambda\| + \Dpu + \Dqu\nonumber\\
    &=(1-\frac{\DTu}{\lambda})\|\wh\T_u\wh\imptheta_\lambda-\wh q_u+\wh p_u\|\nonumber\\
    &\quad\quad\quad\quad +\frac{\DTu}{\lambda}\left(\|\wh\T_u\wh\imptheta_\lambda-\wh q_u+\wh p_u\| +\lambda \|\wh\imptheta_\lambda\| \right) \nonumber\\
    &\quad\quad\quad\quad\quad\quad\quad\quad+\Dpu + \Dqu\nonumber\\
    &\leq \max\{(1-\frac{\DTu}{\lambda}),0\}\|\wh\T_u\wh\imptheta_\lambda-\wh q_u+\wh p_u\| \nonumber\\
    &\quad\quad\quad\quad+\frac{\DTu}{\lambda}\min_{\imptheta'}\left(\|\wh\T_u\imptheta'-\wh q_u+\wh p_u\| + 
    \lambda\|\imptheta'\| \right)\nonumber\\
    &\quad\quad\quad\quad\quad\quad\quad\quad+ \Dpu + \Dqu.
\end{align}
Now using Eq.~\ref{eq:lambdaupperbund}, we have,
\begin{align}
    \|\T_u\wh\imptheta_\lambda&-q_u+p_u\|\nonumber\\
    &\leq \max\{(1-\frac{\DTu}{\lambda}),0\}\left(\min_{\imptheta'}\|\wh \T_u\imptheta'-\wh q_u+\wh p_u\|+\lambda\|\imptheta'\|\right)\nonumber\\
    &\quad\quad\quad\quad+\frac{\DTu}{\lambda}\min_{\imptheta'}\left(\|\wh\T_u\imptheta'-\wh q_u+\wh p_u\| + \lambda\|\imptheta'\| \right)\nonumber\\
    &\quad\quad\quad\quad\quad\quad\quad\quad+ \Dpu + \Dqu\nonumber\\
    &= \max\{1,\frac{\DTu}{\lambda}\}\left(\min_{\imptheta'}\|\wh \T_u\imptheta'-\wh q_u+\wh p_u\|+\lambda\|\imptheta'\|\right) \nonumber\\
    &\quad\quad\quad\quad\quad\quad\quad\quad+ \Dpu + \Dqu.
\end{align}
Applying other side of Lemma~\ref{lemma:difference} for $\wh\imptheta_\lambda$, we have;
\begin{align}
    \|\T_u\wh\imptheta_\lambda&-q_u+p_u\|\nonumber\\
    &\leq\max\{1,\frac{\DTu}{\lambda}\}\left(\min_{\imptheta'}\|\wh \T_u\imptheta'-\wh q_u+\wh p_u\|+\lambda\|\imptheta'\|\right) \nonumber\\
    &\quad\quad\quad\quad\quad\quad\quad\quad+ \Dpu + \Dqu\nonumber\\
    &\leq\max\{1,\frac{\DTu}{\lambda}\}\Big(\min_{\imptheta'}\| \T_u\imptheta'- q_u+ p_u\|\nonumber\\
    &\quad\quad\quad\quad+\DTu\|\imptheta'\|+\Dpu + \Dqu+\lambda\|\imptheta'\|\Big) \nonumber\\
    &\quad\quad\quad\quad\quad\quad\quad\quad+ \Dpu + \Dqu\nonumber\\
    &=\max\{1,\frac{\DTu}{\lambda}\}\Big(\min_{\imptheta'}\| \T_u\imptheta'- q_u+ p_u\|\nonumber\\
    &\quad\quad\quad\quad+(\DTu+\lambda)\|\imptheta'\|\Big) \nonumber\\
    &\quad\quad\quad\quad\quad\quad\quad\quad+ \max\{2,1+\frac{\DTu}{\lambda}\}(\Dpu + \Dqu),
\end{align}
which is the statement of the theorem. 
\end{proof}

\begin{proof}[Lemma~\ref{lemma:regularizedVFunc}]
We directly apply the statement of the Lemma \ref{lemma:lambda} to derive the statement of this Lemma~\ref{lemma:regularizedVFunc}.$\wh\imptheta_\lambda$, the solution to Eq.~\ref{eq:lambda}, and $\wh\imptheta$, the solution to Eq.~\ref{eq:regFun}, are equal when we set $\lambda$ to $\DTu$. Therefore, using Lemma~\ref{lemma:lambda}, we get

\begin{align}
    \|\T_u\wh\imptheta&-q_u+p_u\|\nonumber\\
    &\quad\quad\quad\quad\leq\left(\min_{\imptheta'}\| \T_u\imptheta'- q_u+ p_u\|+2\DTu\|\imptheta'\|\right) \nonumber\\
    &\quad\quad\quad\quad\quad\quad\quad\quad+ 2(\Dpu + \Dqu).
\end{align}
Using the fact that, for the true $\imptheta$, we have that $\| \T_u\imptheta- q_u+ p_u\|=0$, we have 
\begin{align}\label{eq:upperboundFunc}
    \|\T_u(\wh\imptheta&-\imptheta)\|\nonumber\\
    &\leq\left(\min_{\imptheta'}\| \T_u\imptheta'- q_u+ p_u\|+2\DTu\|\imptheta'\|\right) \nonumber\\
    &\quad\quad\quad\quad\quad\quad\quad\quad+ 2(\Dpu + \Dqu),
\end{align}
which states the first statement in the Lemma~\ref{lemma:regularizedVFunc}. Again using the fact that for the true $\imptheta$, we have that $\| \T_u\imptheta- q_u+ p_u\|=0$, plugging in the true $\imptheta$ on the right hand side of Eq.~\ref{eq:upperboundFunc}, we have
\begin{align}
    \|\T_u(\wh\imptheta-\imptheta)\|\leq&2\DTu\|\imptheta\|+ 2(\Dpu + \Dqu).
\end{align}
Using the above statement, we have;
\begin{align}
    \|(\wh\imptheta-\imptheta)\|\leq&2\|\T_u^{-1}\|(\DTu\|\imptheta\|+ \Dpu + \Dqu),
\end{align}
which state the second statement in the Lemma~\ref{lemma:regularizedVFunc}.
\end{proof}

\section{Neural Operator}\label{apx:neural_operator}
In this section, we provide  a discussion on how one may use neural networks to approximate operators. 

\textit{Disclaimer: The following study is for the sake of discussion. We did not attempt to make the results tight and did not attempt to make them general. A further significantly involved study is required to generalize the following results}.

This discussion is motivated by series of works on neural operators where the kernel is learned using a deep neural network and  Nystr\"om approximation~\citep{nystrom1930praktische} is deployed to approximate the integral~\citep{li2020fourier,li2020neural,li2020multipole}. 

We provide this discussion for a general case of integral operator in $\HS$ spaces induced by a symmetric positive definite continuous reproducing kernel $\kernel:\Y\times\Y\rightarrow\Re$, such that $\barkernel:=\sup_{y\in\Y}\kernel(y,y)$ is finite. Consider an integral operator $\T:\Hilbert\rightarrow\Hilbert$ such that,
\begin{align*}
    \T:=\int_\Y(\kernel_y\otimes\kernel_y)d\mu(y),
\end{align*}
for a measure $\mu$. We assume $\mu$ is finite measure. For simplicity. Note that, for any function in $f\in\Hilbert$, and $y'\in\Y$, we have $\T f(y'):=\int_\Y\kernel(y',y)f(y)d\mu(y)$. 

Since $\kernel(\cdot,\cdot)$ is continuous, then, for the compact space $\Y\times\Y\in\Re^d\times\Re^d$, using the universal approximation results of \citet{cybenko1989approximation} for neural networks with non-polynomial activation, for any $\iota>0$, there exists a neural network $\varkappa_\iota$, a continuous function, such that,
\begin{align*}
    \sup_{(y',y)\in\Y\times\Y}|\kernel(y',y)-\varkappa_\iota(y',y)|\leq \iota
\end{align*}
We define a difference function $h:=\varkappa_\iota-\kernel\in \L^\infty$.
Using this results, we derive the deviations in the induced operators, $\T$, and $\T_{\varkappa_\iota}$, where $\T_{\varkappa_\iota}$ is such that for any $f\in\Hilbert$, we have,
\begin{align*}
\T_{\varkappa_\iota} f(y'):=\int_\Y\varkappa_\iota(y',y)f(y)d\mu(y),    
\end{align*}
which exists for the finite measure $\mu$ if   $\T f$ exists.

Our first results elaborates  on the approximation error in $\T_{\varkappa_\iota}f-\T f$. 

\begin{proposition}\label{prop1}
For any $f\in\Hilbert\cap \L^1(\mu)$, under the above construction, we have, \begin{align}
    \|\T_{\varkappa_\iota}f-\T f\|_{\L^\infty} \leq \iota \|f\|_{\L^1(\mu)}
\end{align}
\end{proposition}
\begin{proof} of Proposition \ref{prop1}

For any $y'\in\Y$, and $f\in\Hilbert\cap \L^1(\mu)$ we have,
\begin{align*}
\T f(y')
-\T_{\varkappa_\iota} f(y'):&=\int_\Y\kernel(y',y)f(y)d\mu(y)\nonumber\\
&\quad\quad\quad\quad-\int_\Y\varkappa_\iota(y',y)f(y)d\mu(y)\\
&=\int_\Y\kernel(y',y)f(y)d\mu(y)\nonumber\\
&\quad\quad\quad\quad-\int_\Y(\kernel+h)(y',y)f(y)d\mu(y)\\
&=\int_\Y h(y',y)f(y)d\mu(y)\\
&=\int_{y:f(y)\geq0} h(y',y)f(y)d\mu(y)\nonumber\\
&\quad\quad\quad\quad+\int_{y:f(y)<0} h(y',y)f(y)d\mu(y)
\\
&\leq\iota \Big(\int_{y:f(y)\geq0}  f(y)d\mu(y)\nonumber\\
&\quad\quad\quad\quad-\int_{y:f(y)<0} f(y)d\mu(y)\Big)\nonumber\\
&=\iota \|f\|_{\L^1(\mu)}
\end{align*}
With a similar argument we have,
\begin{align*}
\T f(y')
-\T_{\varkappa_\iota} f(y')
&\geq\iota \Big(-\int_{y:f(y)\geq0}  f(y)d\mu(y)\nonumber\\
&\quad\quad\quad\quad+\int_{y:f(y)<0} f(y)d\mu(y)\Big)\nonumber\\
&=-\iota \|f\|_{\L^1(\mu)}
\end{align*}
Putting these two together results in the final statement. 

\end{proof}

The result in the Proposition \ref{prop1} states that for any function in $\Hilbert\cap \L^1(\mu)$, we can expect the result of neural operator $\T_{\varkappa_\iota}$ is close to that of $\T$. However, this results does not provide approximation in the space of operators. One might be interested in the closeness of $\T_{\varkappa_\iota}$ and $\T$ in some sense. 


Consider the function space $\L^2(\mu)$, and a countable set of its bases functions $\{e_i\}_i$. Also, for the product measure $\mu\times\mu$, let $\{\phi_{ij}:=e_i\times e_j\}_i$ denote the corresponding set of basis for $\L^2(\mu\times\mu)$. Note that the set $\{\phi_{ij}\}_{i,j}$ is not required to be complete.

\begin{proposition}\label{prop2}
Under the above construction, for any countable set $\{e_i\}_i$, we have,
\begin{align*}
\sum_{i}\|(\T
-\T_{\varkappa_\iota})e_i\|_{\L^2(\mu)}^2\leq  \int_{\Y\times\Y}\iota^2d(\mu\times\mu)
\end{align*}
\end{proposition}
\begin{proof} of Proposition~\ref{prop2}.

For any $i,j\in\Natural$, we have
\begin{align*}
    \langle h,\phi_{i,j} 
    \rangle &= \int_{\Y\times\Y}h(y',y)\phi(y',y)d(\mu\times\mu)(y',y)\\
    &=\int_{\Y\times\Y}h(y',y)e_i(y')e(y)d(\mu\times\mu)(y',y)\\
    &=\int_{\Y\times\Y}(\varkappa(y',y)-\kernel(y',y))e_i(y')e(y)d(\mu\times\mu)(y',y)\\
    &= \langle(\T
-\T_{\varkappa_\iota})e_i,e_j\rangle_{\L^2(\mu)}
\end{align*}
Note that, since $h\in\L^\infty$, it is also in $\L^2$. Therefore, $\|h\|^2_{\L^2}\leq \int_{\Y\times\Y}\iota^2d(\mu\times\mu)$. On the other hand, we have $ \sum_{i,j}\langle h,\phi_{ij}\rangle^2_{\L^2(\mu)}\leq\|h\|^2_{\L^2}$ since we did not require $\{\phi_{ij}\}_{i,j}$ to form a compete bases. Putting these statements together, we have,
\begin{align*}
 \sum_{i,j}\langle h,\phi_{ij}\rangle^2_{\L^2(\mu)} &=\sum_{i,j}\langle(\T
-\T_{\varkappa_\iota})e_i,e_j\rangle_{\L^2(\mu)}^2\nonumber\\
&=\sum_{i}\|(\T
-\T_{\varkappa_\iota})e_i\|_{\L^2(\mu)}^2
\end{align*}
Ergo, we have,
\begin{align*}
\sum_{i}\|(\T
-\T_{\varkappa_\iota})e_i\|_{\L^2(\mu)}^2&\leq \|h\|^2_{\L^2}\nonumber\\
&\leq  \int_{\Y\times\Y}\iota^2d(\mu\times\mu)
\end{align*}

\end{proof}

The result of the Proposition~\ref{prop2} states that, a neural network can be used to construct a neural operator that approximates well a class of $\HS$ integral operators in $\L^2(\mu)$ sense.

This study motivates that one can deploy neural networks to tackle optimizations induced in Eqs.~\ref{eq:E3} and \ref{eq:regFun}.

\section{Details in the Experimental Study}\label{Apx:exp}

In the first part of the experiment, we study the setting where $\Y$ is a finite set. For this experiment, we have $\X\in\Re$, and 
we use multilayered neural network, with the size of hidden layer $(50,200,500,200,50)$ for $g$. When $g$ is a classifier, we have an extra softmax layer in the end. We use sklearn package to train these models. In particular, we use \\

MLPRegressor(solver='lbfgs', alpha=1e-1,

\quad learning\_rate = 'adaptive', learning\_rate\_init= 1e-3 ,

\quad max\_iter=5000, activation='relu',

\quad hidden\_layer\_sizes=(50, 200, 500, 200, 50))\\

When $g$ has a softmax layer in the last layer, i.e., its output is on the simplex, we use logistic regression loss to train it. When there is no softmax layer, we use one hot encoding representation of labels, and use L2 loss for training.

To generate the data, we first set the probability of labels as follows,\\

        Py = np.zeros(nc) ; Py[0:int(nc):2] = 1/nc ; Py[1:int(nc):2] = 3/nc  ; Py = Py/np.sum(np.abs(Py))\\

        Qy = np.zeros(nc) ; Qy[0:int(nc):2] = 3/nc  ; Qy[1:int(nc):2] = 1/nc ; Qy = Qy/np.sum(np.abs(Qy))

where nc is the number of classes. We first draw samples for labels, and for each label, we draw a Gaussian random variable, and set x to be the label with an additive Gaussian noise.

The relative error is the L2 norm of error divided by the L2 norm of the importance weight vector. 

For the case of regression, we have $\X\in\Re$ and $\Y\in\Re$. We use Gaussian process regression methods for $u$. In particular, we use 

K\_rbf = RBF(length\_scale=.9, length\_scale\_bounds=(1e-2, 1e3))

and \\

kernel = 1.0 * K\_rbf 
    + WhiteKernel(noise\_level=0.01, 
    
    \quad \quad noise\_level\_bounds=(1e-10, 1e+1))
    
for the choice of kernel, and use \\

GaussianProcessRegressor(kernel=kernel,alpha=0.0)

for the regression. 

The kernel used in order to estimate $\theta$ is $K\_{rbf}$. For estimation, we use the ralxed objective of

\begin{align*}
    \min_{\wh\imptheta'}\|\wh \T_u\imptheta'-\wh q_u+\wh p_u\|^2+\DTu\|\imptheta'\|^2
\end{align*}

Using a similar argument as in representer Theorem, for $\wh\imptheta$, the minimizer of the above optimization, has the following form,

\begin{align*}
    \wh\imptheta = \sum_i^{\alpha n}\beta_i\kernel(y_i)
\end{align*}

Therefore, we are left with estimating $\{\beta_i\}$ for which we deploy the kernel machinery to estimate efficiently.  Please refer to the code for the implementation.

For this setting, we draw samples by first drawing samples for $y$'s. In this case, we draw samples from a distribution with PDF $1-a+2ay$ for the source, and $1-b+2by$ for the target for $y\in[0,1]$ and $a,b\in(0,1)$. Note that, in this case, the importance weight function is $\frac{2by+1-b}{2ay+1-a}$. We set x to be y with an additive Gaussian noise. 

Since the estimator is a function, in order to compute the relative error, we use a grid of 100 points in the interval of $[0,1]$. The relative error is the L2 norm of the error computed on these points, divided by the L2 norm of the true importance weight function computed on the same grid.

\end{document}